\documentclass[10pt,journal,compsoc]{IEEEtran}

\ifCLASSOPTIONcompsoc
  \usepackage[nocompress]{cite}
\else
  \usepackage{cite}
\fi
\ifCLASSINFOpdf
\else
\fi

\usepackage[utf8]{inputenc}

\usepackage[dvipsnames]{xcolor}
\usepackage{cite}
\usepackage{graphicx}
\usepackage{amsmath}
\usepackage{algorithmic}
\usepackage{relsize}
\usepackage{tikz}
\usepackage{pgfplots}
\usepackage{acronym}
\usepackage{tikz}
\usetikzlibrary{shapes,arrows}
\usepackage{pgfplots}
\usepackage{pgfplotstable}
\usepackage{tabularx}
\usepackage{booktabs}
\usepackage{url}
\usepackage{subcaption}

\usepackage{pifont}
\newcommand{\cmark}{\ding{51}}
\newcommand{\xmark}{\ding{55}}

\usepackage{xargs}
\usepackage[colorinlistoftodos,prependcaption]{todonotes}
\newcommandx{\unsure}[2][1=]{\todo[linecolor=red,backgroundcolor=red!25,bordercolor=red,#1]{#2}}
\newcommandx{\change}[2][1=]{\todo[linecolor=blue,backgroundcolor=blue!25,bordercolor=blue,#1]{#2}}
\newcommandx{\info}[2][1=]{\todo[linecolor=OliveGreen,backgroundcolor=OliveGreen!25,bordercolor=OliveGreen,#1]{#2}}
\newcommandx{\improvement}[2][1=]{\todo[inline, linecolor=Plum,backgroundcolor=Plum!25,bordercolor=Plum,#1]{#2}}
\newcommandx{\thiswillnotshow}[2][1=]{\todo[disable,#1]{#2}}

\acrodef{2D}{two-dimensional}
\acrodef{2.5D}{two-and-a-half-dimensional}
\acrodef{3D}{three-dimensional}
\acrodef{C3D}{Convolutional 3D}
\acrodef{CAD}{Computer Aided Design}
\acrodef{CNN}{Convolutional Neural Network}
\acrodef{COCO}{Common Objects in Context}
\acrodef{CRF}{Conditional Random Field}
\acrodef{DAVIS}{Densely-Annotated VIdeo Segmentation}
\acrodef{ECCV}{European Conference on Computer Vision}
\acrodef{FCN}{Fully Convolutional Network}
\acrodef{FWIoU}{Frequency Weighted Intersection over Union}
\acrodef{GCN}{Graph Convolutional Network}
\acrodef{GPU}{Graphics Processing Unit}
\acrodef{ILSVRC}{ImageNet Large Scale Visual Recognition Challenge}
\acrodef{HHA}{Horizontal Height Angle}
\acrodef{IoU}{Intersection over Union}
\acrodef{MCG}{Multi-scale COmbinatorial Grouping}
\acrodef{MINC}{Materials in Context}
\acrodef{MIoU}{Mean Intersection over Union}
\acrodef{MLP}{Multi Layer Perceptron}
\acrodef{MPA}{Mean Pixel Accuracy}
\acrodef{NiN}{Network in Network}
\acrodef{NMS}{Non-Maximum Suppression}
\acrodef{PA}{Pixel Accuracy}
\acrodef{VGG}{Visual Geometry Group}
\acrodef{VOC}{Visual Object Classes}
\acrodef{R-CNN}{Region-CNN}
\acrodef{ReLU}{Rectified Linear Unit}
\acrodef{RGB}{Red Green Blue}
\acrodef{RGB-D}{RGB-Depth}
\acrodef{RNN}{Recurrent Neural Network}
\acrodef{SBD}{Semantic Boundaries Dataset}
\acrodef{SDS}{Simultaneous Detection and Segmentation}
\acrodef{SVM}{Support Vector Machine}
\acrodef{SYNTHIA}{SYNTHetic Collection of Imagery and Annotations}
\acrodef{MDRNN}{Multi-dimensional Recurrent Neural Network}
\acrodef{LSTM-CF}{Long Short-Term Memorized Context Fusion}
\acrodef{LSTM}{Long Short-Term Memory}
\acrodef{GRU}{Gated Recurrent Unit}
\acrodef{rCNN}{Recurrent Convolutional Neural Network}

\begin{document}
	
\tikzstyle{block} = [draw, rectangle, minimum height=3em, minimum width=6em]
\tikzstyle{edge} = [draw, ->, line width= 0.1em]
\tikzstyle{sum} = [draw, circle, node distance=1.5cm]
\tikzstyle{input} = [coordinate]
\tikzstyle{output} = [coordinate]
\tikzstyle{pinstyle} = [pin edge={to-,thin,black}]

%
\title{A Review on Deep Learning Techniques\\ Applied to Semantic Segmentation}
%
%
%
%

\author{A.~Garcia-Garcia,
        S.~Orts-Escolano,
				S.O.~Oprea,
				V.~Villena-Martinez,
        and~J.~Garcia-Rodriguez
\IEEEcompsocitemizethanks{\IEEEcompsocthanksitem A. Garcia-Garcia, S.O. Oprea, V. Villena-Martinez, and J. Garcia-Rodriguez  are with the Department
of Computer Technology, University of Alicante, Spain. \protect\\
E-mail: \{agarcia, soprea, vvillena, jgarcia\}@dtic.ua.es
\IEEEcompsocthanksitem S. Orts-Escolano is with the Department of Computer Science and Artificial Intelligence, Universit of Alicante, Spain. \protect\\
E-mail: sorts@ua.es.}
}
\IEEEtitleabstractindextext{%
\begin{abstract}
Image semantic segmentation is more and more being of interest for computer vision and machine learning researchers. Many applications on the rise need accurate and efficient segmentation mechanisms: autonomous driving, indoor navigation, and even virtual or augmented reality systems to name a few. This demand coincides with the rise of deep learning approaches in almost every field or application target related to computer vision, including semantic segmentation or scene understanding. This paper provides a review on deep learning methods for semantic segmentation applied to various application areas. Firstly, we describe the terminology of this field as well as mandatory background concepts. Next, the main datasets and challenges are exposed to help researchers decide which are the ones that best suit their needs and their targets. Then, existing methods are reviewed, highlighting their contributions and their significance in the field. Finally, quantitative results are given for the described methods and the datasets in which they were evaluated, following up with a discussion of the results. At last, we point out a set of promising future works and draw our own conclusions about the state of the art of semantic segmentation using deep learning techniques.
\end{abstract}

}

\maketitle

\IEEEdisplaynontitleabstractindextext

%
\IEEEpeerreviewmaketitle

\ifCLASSOPTIONcompsoc
\IEEEraisesectionheading{\section{Introduction}\label{sec:introduction}}
\else
\section{Introduction}
\label{sec:introduction}
\fi

\IEEEPARstart{N}{owadays}, semantic segmentation -- applied to still 2D images, video, and even 3D or volumetric data -- is one of the key problems in the field of computer vision. Looking at the big picture, semantic segmentation is one of the high-level task that paves the way towards complete scene understanding. The importance of scene understanding as a core computer vision problem is highlighted by the fact that an increasing number of applications nourish from inferring knowledge from imagery. Some of those applications include autonomous driving \cite{Ess2009}\cite{Geiger2012}\cite{Cordts2016}, human-machine interaction \cite{Oberweger2015}, computational photography \cite{Yoon2015}, image search engines \cite{Wan2014}, and augmented reality to name a few. Such problem has been addressed in the past using various traditional computer vision and machine learning techniques. Despite the popularity of those kind of methods, the deep learning revolution has turned the tables so that many computer vision problems -- semantic segmentation among them -- are being tackled using deep architectures, usually \acp{CNN} \cite{Ning2005}\cite{Ciresan2012}\cite{Farabet2013}\cite{Hariharan2014}\cite{Gupta2014}, which are surpassing other approaches by a large margin in terms of accuracy and sometimes even efficiency. However, deep learning is far from the maturity achieved by other old-established branches of computer vision and machine learning. Because of that, there is a lack of unifying works and state of the art reviews. The ever-changing state of the field makes initiation difficult and keeping up with its evolution pace is an incredibly time-consuming task due to the sheer amount of new literature being produced. This makes it hard to keep track of the works dealing with semantic segmentation and properly interpret their proposals, prune subpar approaches, and validate results.

\begin{figure}[!b]
	\centering
	\includegraphics[width=0.8\linewidth]{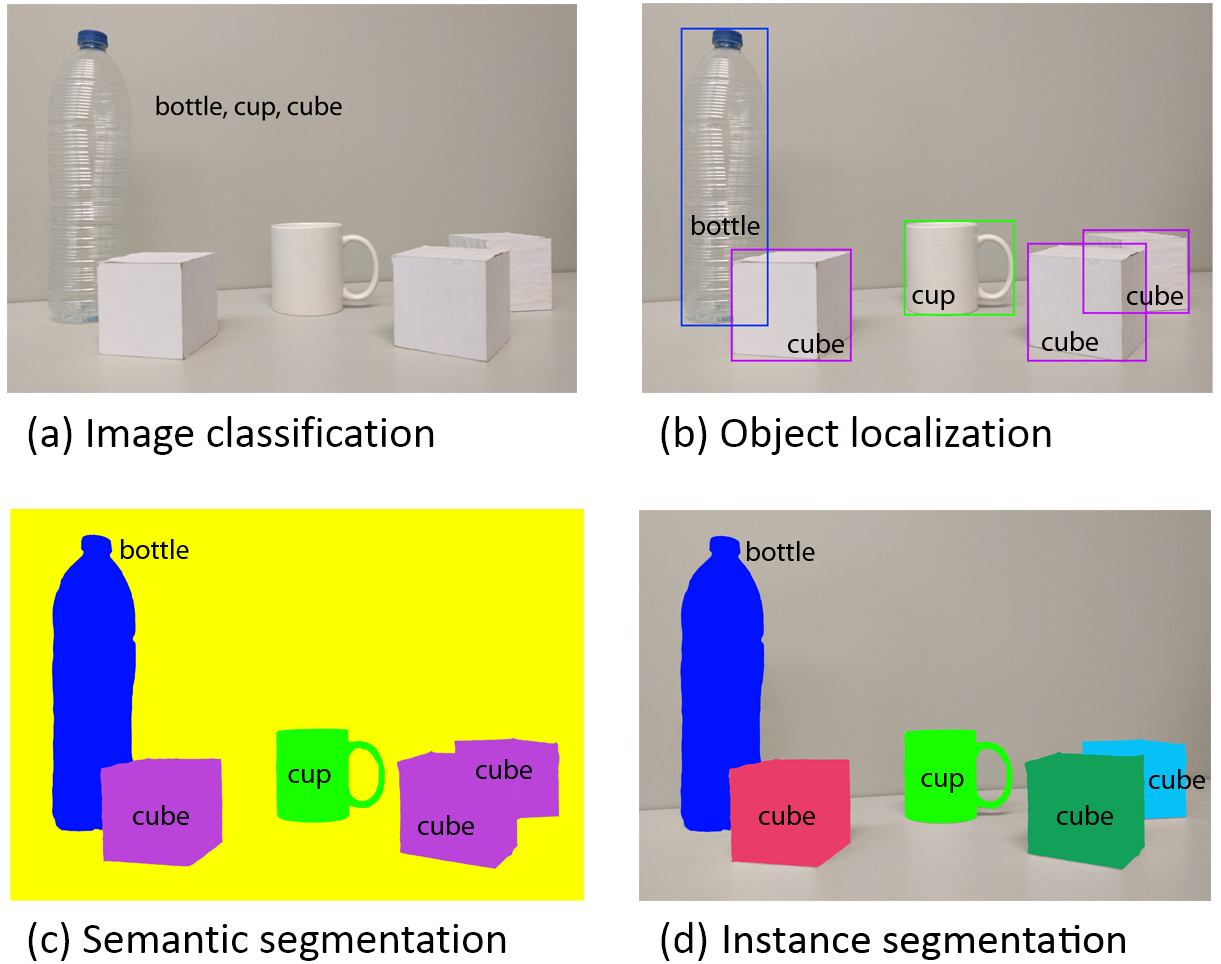}
	\caption{Evolution of object recognition or scene understanding from coarse-grained to fine-grained inference: classification, detection or localization, semantic segmentation, and instance segmentation.}
	\label{fig:background_evolution}
\end{figure}

To the best of our knowledge, this is the first review to focus explicitly on deep learning for semantic segmentation. Various semantic segmentation surveys already exist such as the works by Zhu \emph{et al.}\cite{Zhu2016} and Thoma\cite{Thoma2016}, which do a great work summarizing and classifying existing methods, discussing datasets and metrics, and providing design choices for future research directions. However, they lack some of the most recent datasets, they do not analyze frameworks, and none of them provide details about deep learning techniques. Because of that, we consider our work to be novel and helpful thus making it a significant contribution for the research community.

\newpage

The key contributions of our work are as follows:

\begin{itemize}
	\item We provide a broad survey of existing datasets that might be useful for segmentation projects with deep learning techniques.
	\item An in-depth and organized review of the most significant methods that use deep learning for semantic segmentation, their origins, and their contributions.
	\item A thorough performance evaluation which gathers quantitative metrics such as accuracy, execution time, and memory footprint.
	\item A discussion about the aforementioned results, as well as a list of possible future works that might set the course of upcoming advances, and a conclusion summarizing the state of the art of the field.
\end{itemize}

The remainder of this paper is organized as follows. Firstly, Section \ref{sec:background} introduces the semantic segmentation problem as well as notation and conventions commonly used in the literature. Other background concepts such as common deep neural networks are also reviewed.  Next, Section \ref{sec:datasets} describes existing datasets, challenges, and benchmarks. Section \ref{sec:methods} reviews existing methods following a bottom-up complexity order based on their contributions. This section focuses on describing the theory and highlights of those methods rather than performing a quantitative evaluation. Finally, Section \ref{sec:discussion} presents a brief discussion on the presented methods based on their quantitative results on the aforementioned datasets. In addition, future research directions are also laid out. At last, Section \ref{sec:conclusion} summarizes the paper and draws conclusions about this work and the state of the art of the field.

\section{Terminology and Background Concepts}
\label{sec:background}

In order to properly understand how semantic segmentation is tackled by modern deep learning architectures, it is important to know that it is not an isolated field but rather a natural step in the progression from coarse to fine inference. The origin could be located at classification, which consists of making a prediction for a whole input, i.e., predicting which is the object in an image or even providing a ranked list if there are many of them. Localization or detection is the next step towards fine-grained inference, providing not only the classes but also additional information regarding the spatial location of those classes, e.g., centroids or bounding boxes. Providing that, it is obvious that semantic segmentation is the natural step to achieve fine-grained inference, its goal: make dense predictions inferring labels for every pixel; this way, each pixel is labeled with the class of its enclosing object or region. Further improvements can be made, such as instance segmentation (separate labels for different instances of the same class) and even part-based segmentation (low-level decomposition of already segmented classes into their components). Figure \ref{fig:background_evolution} shows the aforementioned evolution. In this review, we will mainly focus on generic scene labeling, i.e., per-pixel class segmentation, but we will also review the most important methods on instance and part-based segmentation.

In the end, the per-pixel labeling problem can be reduced to the following formulation: find a way to assign a state from the label space $\mathcal{L} = \{l_1, l_2, ... , l_k\}$ to each one of the elements of a set of random variables $\mathcal{X} = \{x_1, x_2, ... , x_N\}$. Each label $l$ represents a different class or object, e.g., aeroplane, car, traffic sign, or background. This label space has $k$ possible states which are usually extended to $k+1$ and treating $l_0$ as background or a void class. Usually, $\mathcal{X}$ is a \acs{2D} image of $W\times H = N$ pixels $x$. However, that set of random variables can be extended to any dimensionality such as volumetric data or hyperspectral images.

Apart from the problem formulation, it is important to remark some background concepts that might help the reader to understand this review. Firstly, common networks, approaches, and design decisions that are often used as the basis for deep semantic segmentation systems. In addition, common techniques for training such as transfer learning. At last, data pre-processing and augmentation approaches.

\subsection{Common Deep Network Architectures}

As we previously stated, certain deep networks have made such significant contributions to the field that they have become widely known standards. It is the case of AlexNet, \acs{VGG}-16, GoogLeNet, and ResNet. Such was their importance that they are currently being used as building blocks for many segmentation architectures. For that reason, we will devote this section to review them.

\subsubsection{AlexNet}

\begin{figure}[!hbt]
	\centering
	\includegraphics[width=\linewidth]{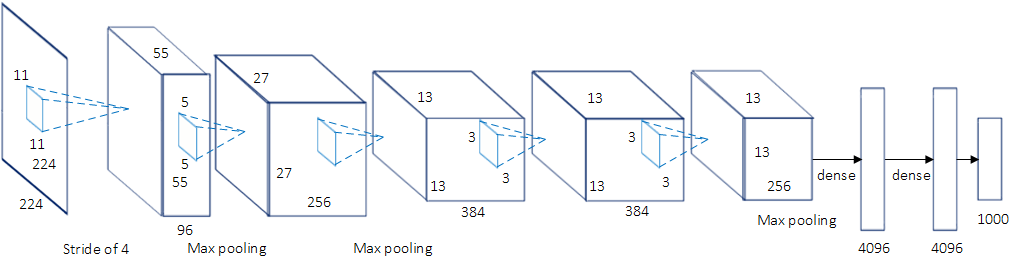}
	\caption{AlexNet \acl{CNN} architecture. Figure reproduced from \cite{Krizhevsky2012}.}
	\label{fig:alexnet}
\end{figure}

AlexNet was the pioneering deep \acs{CNN} that won the \acs{ILSVRC}-2012 with a TOP-5 test accuracy of $84.6\%$ while the closest competitor, which made use of traditional techniques instead of deep architectures, achieved a $73.8\%$ accuracy in the same challenge. The architecture presented by Krizhevsky \emph{et al.} \cite{Krizhevsky2012} was relatively simple. It consists of five convolutional layers, max-pooling ones, \acp{ReLU} as non-linearities, three fully-connected layers, and dropout. Figure \ref{fig:alexnet} shows that \acs{CNN} architecture.

\subsubsection{\acs{VGG}}

\begin{figure}[!hbt]
	\centering
	\includegraphics[width=\linewidth]{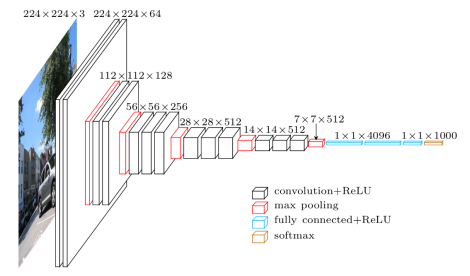}
	\caption{\acs{VGG}-16 \acs{CNN} architecture. Figure extracted from Matthieu Cord's talk with his permission.}
	\label{fig:vgg16}
\end{figure}

\ac{VGG} is a \ac{CNN} model introduced by the \acf{VGG} from the University of Oxford. They proposed various models and configurations of deep \acp{CNN} \cite{Simonyan2014}, one of them was submitted to the \ac{ILSVRC}-2013. That model, also known as \acs{VGG}-16 due to the fact that it is composed by $16$ weight layers, became popular thanks to its achievement of $92.7\%$ TOP-5 test accuracy. Figure \ref{fig:vgg16} shows the configuration of \acs{VGG}-16. The main difference between \acs{VGG}-16 and its predecessors is the use of a stack of convolution layers with small receptive fields in the first layers instead of few layers with big receptive fields. This leads to less parameters and more non-linearities in between, thus making the decision function more discriminative and the model easier to train.

\subsubsection{GoogLeNet}

\begin{figure}[!hbt]
\centering
 \scalebox{.7}
 {
	\begin{tikzpicture}[auto, >=latex']
	\node [block, fill=white] (filterconcat) [align=center] {Filter\\concatenation};
	\node [block, fill=cyan!35, below left of= filterconcat, node distance = 6em] (3x3conv) {3x3 convolutions};
	\node [block, fill=cyan!70, below right of= filterconcat, node distance = 6em] (5x5conv) {5x5 convolutions} ;
	\node [block, fill=cyan!15, left of = 3x3conv, node distance= 10em] (1x1conv0) {1x1 convolutions};
	\node [block, fill=cyan!15, right of= 5x5conv, node distance= 10em] (1x1conv1) {1x1 convolutions};
	\node [block, fill=cyan!15, below of= 3x3conv, node distance=4 em] (1x1conv2) {1x1 convolutions};
	\node [block, fill=cyan!15, below of= 5x5conv, node distance=4 em] (1x1conv3) {1x1 convolutions};
	\node [block, fill=red!40, below of= 1x1conv1, node distance= 4em] (3x3max) {3x3 max pooling};
	\node [block, fill=white, below left of= 1x1conv3, node distance=6em] (prevlayer) {Previous layer};
		
	\draw[edge] (3x3conv.north) |- (filterconcat.west);
	\draw[edge] (1x1conv0.north) |- (filterconcat.west);
	\draw[edge] (5x5conv.north) |- (filterconcat.east);
	\draw[edge] (1x1conv1.north) |- (filterconcat.east);
	\draw[edge] (1x1conv2.north) -- (3x3conv.south);
	\draw[edge] (1x1conv3.north) -- (5x5conv.south);
	\draw[edge] (3x3max.north) -- (1x1conv1.south);
	\draw[edge] (prevlayer) -| (1x1conv0.south);
	\draw[edge] (prevlayer) -| (1x1conv2.south);
	\draw[edge] (prevlayer) -| (1x1conv3.south);
	\draw[edge] (prevlayer) -| (3x3max.south);
	\end{tikzpicture}
}
	\caption{Inception module with dimensionality reduction from the GoogLeNet architecture. Figure reproduced from \cite{Szegedy2015}.}
	\label{fig:inception-module}
\end{figure}
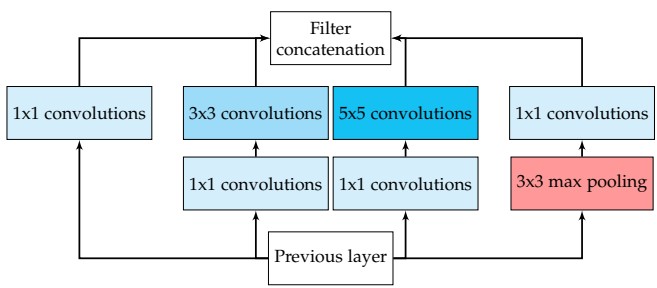

GoogLeNet is a network introduced by Szegedy \emph{et al.} \cite{Szegedy2015} which won the \ac{ILSVRC}-2014 challenge with a TOP-5 test accuracy of $93.3\%$. This \acs{CNN} architecture is characterized by its complexity, emphasized by the fact that it is composed by $22$ layers and a newly introduced building block called \emph{inception} module (see Figure \ref{fig:inception-module}). This new approach proved that \acs{CNN} layers could be stacked in more ways than a typical sequential manner. In fact, those modules consist of a \ac{NiN} layer, a pooling operation, a large-sized convolution layer, and small-sized convolution layer. All of them are computed in parallel and followed by $1\times1$ convolution operations to reduce dimensionality. Thanks to those modules, this network puts special consideration on memory and computational cost by significantly reducing the number of parameters and operations.

\subsubsection{ResNet}

\begin{figure}[!hbt]
	\centering
	\begin{tikzpicture}[auto, node distance=1.8 cm,>=latex']
	\node [input, name=input] {};
	\node [block, fill=cyan!20, below of= input] (weightlayer1) {weight layer};
	\node [block, fill=cyan!20, below of= weightlayer1] (weightlayer2) {weight layer};
	\node [sum, below of= weightlayer2, label={[xshift=-1.2cm, yshift= -0.55cm] $F(\chi) + \chi$}] (sum) {+};
	\coordinate[right of= sum] (midpoint) {};

	\draw[draw, ->] (input) [left] -- node [name=inputedge] {$\chi$} (weightlayer1);
	\draw[->] (weightlayer1) [left] -- node [xshift=-1cm] {$F(\chi)$} (weightlayer2);
	\draw[->] (weightlayer2) -- (sum);
	\draw[-] (inputedge) -| node [yshift=-2cm, align=center] {$\chi$\\identity}(midpoint);
	\draw[->] (midpoint) -- node {$relu$} (sum.east);
	\end{tikzpicture}
	\caption{Residual block from the ResNet architecture. Figure reproduced from \cite{He2016}.}
	\label{fig:resnet-block}
\end{figure}
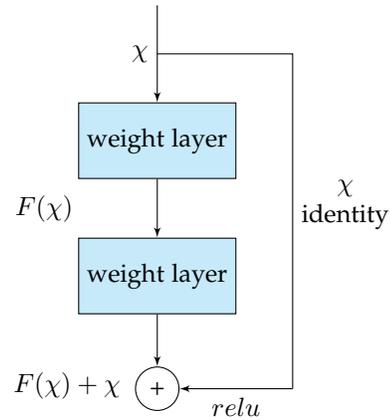

Microsoft's ResNet\cite{He2016} is specially remarkable thanks to winning \ac{ILSVRC}-2016 with $96.4\%$ accuracy. Apart from that fact, the network is well-known due to its depth ($152$ layers) and the introduction of residual blocks (see Figure \ref{fig:resnet-block}). The residual blocks address the problem of training a really deep architecture by introducing identity skip connections so that layers can copy their inputs to the next layer.

The intuitive idea behind this approach is that it ensures that the next layer learns something new and different from what the input has already encoded (since it is provided with both the output of the previous layer and its unchanged input). In addition, this kind of connections help overcoming the vanishing gradients problem.

\subsubsection{ReNet}

\begin{figure}[!hbt]
	\centering
	\includegraphics[width=0.5\linewidth]{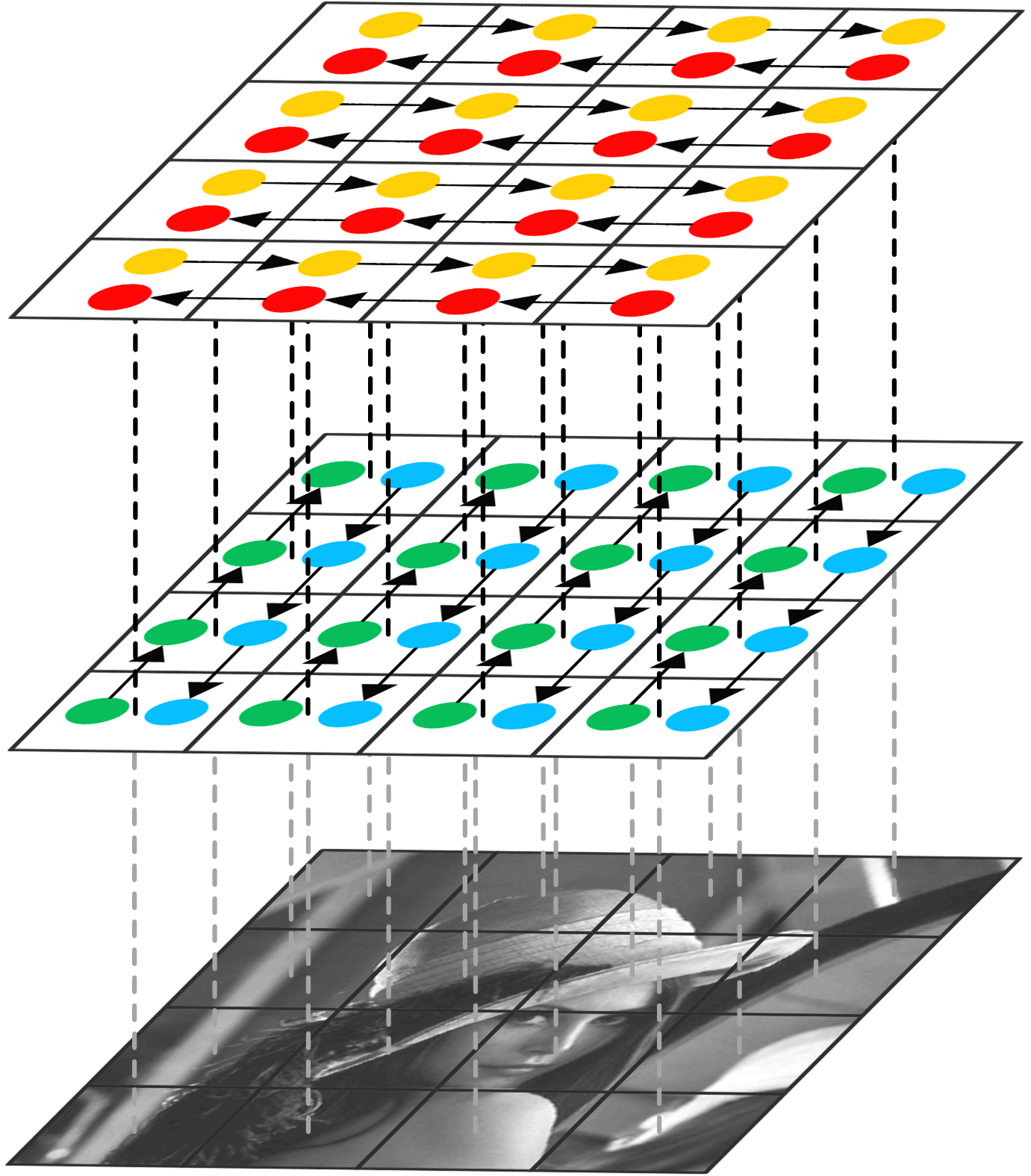}
	\caption{One layer of ReNet architecture modeling vertical and horizontal spatial dependencies. Extracted from \cite{Visin2015}.}
	\label{fig:renet}
\end{figure}

In order to extend \acp{RNN} architectures to multi-dimensional tasks, Graves et al. \cite{Graves2007} proposed a \ac{MDRNN} architecture which replaces each single recurrent connection from standard \acsp{RNN} with $d$ connections, where $d$ is the number of spatio-temporal data dimensions. Based on this initial approach, Visin el al. \cite{Visin2015} proposed ReNet architecture in which instead of multidimensional \acsp{RNN}, they have been using usual sequence \acsp{RNN}. In this way, the number of \acsp{RNN} is scaling linearly at each layer regarding to the number of dimensions $d$ of the input image ($2d$). In this approach, each convolutional layer (convolution + pooling) is replaced with four \acsp{RNN} sweeping the image vertically and horizontally in both directions as we can see in Figure \ref{fig:renet}.

\subsection{Transfer Learning}

Training a deep neural network from scratch is often not feasible because of various reasons: a dataset of sufficient size is required (and not usually available) and reaching convergence can take too long for the experiments to be worth. Even if a dataset large enough is available and convergence does not take that long, it is often helpful to start with pre-trained weights instead of random initialized ones\cite{Ahmed2008}\cite{Oquab2014}. Fine-tuning the weights of a pre-trained network by continuing with the training process is one of the major transfer learning scenarios.

Yosinski \emph{et al.}\cite{Yosinski2014} proved that transferring features even from distant tasks can be better than using random initialization, taking into account that the transferability of features decreases as the difference between the pre-trained task and the target one increases.

However, applying this transfer learning technique is not completely straightforward. On the one hand, there are architectural constraints that must be met to use a pre-trained network. Nevertheless, since it is not usual to come up with a whole new architecture, it is common to reuse already existing network architectures (or components) thus enabling transfer learning. On the other hand, the training process differs slightly when fine-tuning instead of training from scratch. It is important to choose properly which layers to fine-tune -- usually the higher-level part of the network, since the lower one tends to contain more generic features -- and also pick an appropriate policy for the learning rate, which is usually smaller due to the fact that the pre-trained weights are expected to be relatively good so there is no need to drastically change them.

Due to the inherent difficulty of gathering and creating per-pixel labelled segmentation datasets, their scale is not as large as the size of classification datasets such as ImageNet\cite{Deng2009}\cite{Russakovsky2015}. This problem gets even worse when dealing with \acs{RGB-D} or \acs{3D} datasets, which are even smaller. For that reason, transfer learning, and in particular fine-tuning from pre-trained classification networks is a common trend for segmentation networks and has been successfully applied in the methods that we will review in the following sections.

\subsection{Data Preprocessing and Augmentation}

Data augmentation is a common technique that has been proven to benefit the training of machine learning models in general and deep architectures in particular; either speeding up convergence or acting as a regularizer, thus avoiding overfitting and increasing generalization capabilities\cite{Wong2016}.

It typically consist of applying a set of transformations in either data or feature spaces, or even both. The most common augmentations are performed in the data space. That kind of augmentation generates new samples by applying transformations to the already existing data. There are many transformations that can be applied: translation, rotation, warping, scaling, color space shifts, crops, etc. The goal of those transformations is to generate more samples to create a larger dataset, preventing overfitting and presumably regularizing the model, balance the classes within that database, and even synthetically produce new samples that are more representative for the use case or task at hand.

Augmentations are specially helpful for small datasets, and have proven their efficacy with a long track of success stories. For instance, in \cite{Shen2016}, a dataset of $1500$ portrait images is augmented synthesizing four new scales ($0.6, 0.8, 1.2, 1.5$), four new rotations ($-45, -22, 22, 45$), and four gamma variations ($0.5, 0.8, 1.2, 1.5$) to generate a new dataset of $19000$ training images. That process allowed them to raise the accuracy of their system for portrait segmentation from $73.09$ to $94.20$ \ac{IoU} when including that augmented dataset for fine-tuning.

\section{Datasets and Challenges}
\label{sec:datasets}

Two kinds of readers are expected for this type of review: either they are initiating themselves in the problem, or either they are experienced enough and they are just looking for the most recent advances made by other researchers in the last few years. Although the second kind is usually aware of two of the most important aspects to know before starting to research in this problem, it is critical for newcomers to get a grasp of what are the top-quality datasets and challenges. Therefore, the purpose of this section is to kickstart novel scientists, providing them with a brief summary of datasets that might suit their needs as well as  data augmentation and preprocessing tips. Nevertheless, it can also be useful for hardened researchers who want to review the fundamentals or maybe discover new information. 

Arguably, data is one of the most -- if not the most -- important part of any machine learning system. When dealing with deep networks, this importance is increased even more. For that reason, gathering adequate data into a dataset is critical for any segmentation system based on deep learning techniques. Gathering and constructing an appropriate dataset, which must have a scale large enough and represent the use case of the system accurately, needs time, domain expertise to select relevant information, and infrastructure to capture that data and transform it to a representation that the system can properly understand and learn. This task, despite the simplicity of its formulation in comparison with sophisticated neural network architecture definitions, is one of the hardest problems to solve in this context. Because of that, the most sensible approach usually means using an existing standard dataset which is representative enough for the domain of the problem. Following this approach has another advantage for the community: standardized datasets enable fair comparisons between systems; in fact, many datasets are part of a challenge which reserves some data -- not provided to developers to test their algorithms -- for a competition in which many methods are tested, generating a fair ranking of methods according to their actual performance without any kind of data cherry-picking.

In the following lines we describe the most popular large-scale datasets currently in use for semantic segmentation. All datasets listed here provide appropriate pixel-wise or point-wise labels. The list is structured into three parts according to the nature of the data: \acs{2D} or plain \acs{RGB} datasets, \acs{2.5D} or \ac{RGB-D} ones, and pure volumetric or \acs{3D} databases. Table \ref{table:datasets} shows a summarized view, gathering all the described datasets and providing useful information such as their purpose, number of classes, data format, and training/validation/testing splits.

\begin{table*}[!t]
	\centering
	\caption{Popular large-scale segmentation datasets.}
	\resizebox{\linewidth}{!}{
	\begin{tabular}{|c|c|c|c|c|c|c|c|c|c|c|}
		\hline
		Name and Reference & Purpose & Year & Classes & Data & Resolution & Sequence & Synthetic/Real & Samples (training) & Samples (validation) & Samples (test)\\
		\hline
		PASCAL \acs{VOC} 2012 Segmentation \cite{Everingham2015} & Generic & 2012 & 21 & 2D & Variable & \xmark & R & $1464$ & $1449$ & Private\\
		PASCAL-Context \cite{Mottaghi2014} & Generic & 2014 & 540 (59) & 2D & Variable & \xmark & R & $10103$ & N/A & $9637$\\
		PASCAL-Part \cite{Chen2014} & Generic-Part & 2014 & 20 & 2D & Variable & \xmark & R & $10103$ & N/A & $9637$\\
		\acs{SBD} \cite{Hariharan2011} & Generic & 2011 & 21 & 2D & Variable & \xmark & R & $8498$ & $2857$ & N/A\\
		Microsoft \acs{COCO} \cite{Lin2014} & Generic & 2014 & +80 & 2D & Variable & \xmark & R & $82783$ & $40504$ & $81434$\\
		\acs{SYNTHIA} \cite{Ros2016} & Urban (Driving)& 2016 & 11 & 2D & $960\times720$ & \xmark & S & $13407$ & N/A & N/A\\
		Cityscapes (fine) \cite{Cordts2015} & Urban & 2015 &30 (8) & 2D & $2048\times1024$ & \cmark & R & $2975$ & $500$ & $1525$\\
		Cityscapes (coarse) \cite{Cordts2015} & Urban & 2015 & 30 (8) & 2D & $2048\times1024$ & \cmark & R & $22973$ & $500$ & N/A\\
		CamVid \cite{Brostow2009} & Urban (Driving) & 2009 & 32 & 2D & $960\times720$ & \cmark & R & $701$ & N/A & N/A\\
		CamVid-Sturgess \cite{Sturgess2009} & Urban (Driving) & 2009 & 11 & 2D & $960\times720$ & \cmark & R & $367$ & $100$ & $233$\\
		KITTI-Layout \cite{Alvarez2012}\cite{Ros2015a} & Urban/Driving & 2012 & 3 & 2D & Variable & \xmark & R & $323$ & N/A & N/A\\
		KITTI-Ros \cite{Ros2015} & Urban/Driving & 2015 & 11 & 2D & Variable & \xmark & R & $170$ & N/A & $46$\\
		KITTI-Zhang \cite{Zhang2015} & Urban/Driving & 2015 & 10 & 2D/3D & $1226\times370$ & \xmark & R & $140$ & N/A & $112$\\
		Stanford background \cite{Gould2009} & Outdoor & 2009 & 8 & 2D & $320\times240$ & \xmark & R & $725$ & N/A & N/A\\
		SiftFlow \cite{Liu2009} & Outdoor & 2011 & 33 & 2D & $256\times256$ & \xmark  & R & $2688$ & N/A & N/A \\
		Youtube-Objects-Jain \cite{Jain2014} & Objects & 2014 & 10 & 2D & $480\times360$ & \cmark & R & $10167$ & N/A & N/A\\
		Adobe's Portrait Segmentation \cite{Shen2016} & Portrait & 2016 & 2 & 2D & $600\times800$ & \xmark & R & $1500$ & $300$ & N/A\\
		\acs{MINC} \cite{Bell2015} & Materials & 2015 & 23 & 2D & Variable & \xmark & R & $7061$ & $2500$ & $5000$\\
		\acs{DAVIS} \cite{Perazzi2016}\cite{Pont-Tuset2017} & Generic & 2016 & 4 & 2D & 480p & \cmark & R & $4219$ & $2023$ & $2180$\\
		NYUDv2 \cite{Silberman2012} & Indoor & 2012 & 40 & 2.5D & $480\times640$ & \xmark & R & $795$ & $654$ & N/A\\
		SUN3D \cite{Xiao2013} & Indoor & 2013 & -- & 2.5D & $640\times480$ & \cmark & R & $19640$ & N/A & N/A\\
		SUNRGBD \cite{Song2015} & Indoor & 2015 & 37 & 2.5D & Variable & \xmark & R & $2666$ & $2619$ & $5050$\\
		RGB-D Object Dataset\cite{lai2011large} & Household objects & 2011 & 51 & 2.5D & $640\times480$ & \cmark & R & $207920$ & N/A & N/A\\
		ShapeNet Part\cite{Yi2016} & Object/Part & 2016 & 16/50 & 3D & N/A & \xmark & S & $31,963$ & N/A & N/A\\
		Stanford 2D-3D-S\cite{Armeni2017} & Indoor & 2017 & 13 & 2D/2.5D/3D & $1080\times1080$ & \cmark & R & $70469$ & N/A & N/A\\
		3D Mesh \cite{Chen:2009:ABF} & Object/Part & 2009 & 19 & 3D & N/A & \xmark & S & 380 & N/A & N/A\\
		Sydney Urban Objects Dataset\cite{quadros2012feature} & Urban (Objects) & 2013 & 26 & 3D & N/A & \xmark & R & $41$ & N/A & N/A\\
		Large-Scale Point Cloud Classification Benchmark\cite{hackel2016contour} & Urban/Nature & 2016 & 8 & 3D & N/A & \xmark  &  R & $15$ & N/A & $15$\\
		
		\hline
	\end{tabular}}
	\label{table:datasets}
\end{table*}

\subsection{2D Datasets}

Throughout the years, semantic segmentation has been mostly focused on \acl{2D} images. For that reason, \acs{2D} datasets are the most abundant ones. In this section we describe the most popular \acs{2D} large-scale datasets for semantic segmentation, considering \acs{2D} any dataset that contains any kind of \acl{2D} representations such as gray-scale or \ac{RGB} images.

\begin{itemize}
	\item \textbf{PASCAL \acf{VOC}}\cite{Everingham2015}\footnote{\url{http://host.robots.ox.ac.uk/pascal/VOC/voc2012/}}: this challenge consists of a ground-truth annotated dataset of images and five different competitions: classification, detection, segmentation, action classification, and person layout. The segmentation one is specially interesting since its goal is to predict the object class of each pixel for each test image. There are 21 classes categorized into vehicles, household, animals, and other: aeroplane, bicycle, boat, bus, car, motorbike, train, bottle, chair, dining table, potted plant, sofa, TV/monitor, bird, cat, cow, dog, horse, sheep, and person. Background is also considered if the pixel does not belong to any of those classes. The dataset is divided into two subsets: training and validation with 1464 and 1449 images respectively. The test set is private for the challenge. This dataset is arguably the most popular for semantic segmentation so almost every remarkable method in the literature is being submitted to its performance evaluation server to validate against their private test set. Methods can be trained either using only the dataset or either using additional information. Furthermore, its leaderboard is public and can be consulted online\footnote{\url{http://host.robots.ox.ac.uk:8080/leaderboard/displaylb.php?challengeid=11&compid=6}}.
	\item \textbf{PASCAL Context}\cite{Mottaghi2014}\footnote{\url{http://www.cs.stanford.edu/~roozbeh/pascal-context/}}: this dataset is an extension of the PASCAL \ac{VOC} 2010 detection challenge which contains pixel-wise labels for all training images ($10103$). It contains a total of $540$ classes -- including the original 20 classes plus background from PASCAL \ac{VOC} segmentation -- divided into three categories (objects, stuff, and hybrids). Despite the large number of categories, only the 59 most frequent are remarkable. Since its classes follow a power law distribution, there are many of them which are too sparse throughout the dataset. In this regard, this subset of 59 classes is usually selected to conduct studies on this dataset, relabeling the rest of them as background.
	\item \textbf{PASCAL Part}\cite{Chen2014}\footnote{\url{http://www.stat.ucla.edu/~xianjie.chen/pascal_part_dataset/pascal_part.html}}: this database is an extension of the PASCAL \ac{VOC} 2010 detection challenge which goes beyond that task to provide per-pixel segmentation masks for each part of the objects (or at least silhouette annotation if the object does not have a consistent set of parts). The original classes of PASCAL \ac{VOC} are kept, but their parts are introduced, e.g., bicycle is now decomposed into back wheel, chain wheel, front wheel, handlebar, headlight, and saddle. It contains labels for all training and validation images from PASCAL \ac{VOC} as well as for the 9637 testing images.
	\item \textbf{\acf{SBD}}\cite{Hariharan2011}\footnote{\url{http://home.bharathh.info/home/sbd}}: this dataset is an extended version of the aforementioned PASCAL \ac{VOC} which provides semantic segmentation ground truth for those images that were not labelled in \ac{VOC}. It contains annotations for $11355$ images from PASCAL \ac{VOC} $2011$. Those annotations provide both category-level and instance-level information, apart from boundaries for each object. Since the images are obtained from the whole PASCAL \ac{VOC} challenge (not only from the segmentation one), the training and validation splits diverge. In fact, \ac{SBD} provides its own training ($8498$ images) and validation ($2857$ images) splits. Due to its increased amount of training data, this dataset is often used as a substitute for PASCAL \ac{VOC} for deep learning.
	\item \textbf{Microsoft \acf{COCO}}\cite{Lin2014}\footnote{\url{http://mscoco.org/}}: is another image recognition, segmentation, and captioning large-scale dataset. It features various challenges, being the detection one the most relevant for this field since one of its parts is focused on segmentation. That challenge, which features more than $80$ classes, provides more than $82783$ images for training, $40504$ for validation, and its test set consist of more than $80000$ images. In particular, the test set is divided into four different subsets or splits: test-dev ($20000$ images) for additional validation, debugging, test-standard ($20000$ images) is the default test data for the competition and the one used to compare state-of-the-art methods, test-challenge ($20000$ images) is the split used for the challenge when submitting to the evaluation server, and test-reserve ($20000$ images) is a split used to protect against possible overfitting in the challenge (if a method is suspected to have made too many submissions or trained on the test data, its results will be compared with the reserve split). Its popularity and importance has ramped up since its appearance thanks to its large scale. In fact, the results of the challenge are presented yearly on a joint workshop at the \ac{ECCV}\footnote{\url{http://image-net.org/challenges/ilsvrc+coco2016}} together with ImageNet's ones.
	\item \textbf{\ac{SYNTHIA}}\cite{Ros2016}\footnote{\url{http://synthia-dataset.net/}}: is a large-scale collection of photo-realistic renderings of a virtual city, semantically segmented, whose purpose is scene understanding in the context of driving or urban scenarios.The dataset provides fine-grained pixel-level annotations for $11$ classes (void, sky, building, road, sidewalk, fence, vegetation, pole, car, sign, pedestrian, and cyclist). It features $13407$ training images from rendered video streams. It is also characterized by its diversity in terms of scenes (towns, cities, highways), dynamic objects, seasons, and weather.
	\item \textbf{Cityscapes}\cite{Cordts2015}\footnote{\url{https://www.cityscapes-dataset.com/}}: is a large-scale database which focuses on semantic understanding of urban street scenes. It provides semantic, instance-wise, and dense pixel annotations for 30 classes grouped into 8 categories (flat surfaces, humans, vehicles, constructions, objects, nature, sky, and void). The dataset consist of around 5000 fine annotated images and 20000 coarse annotated ones. Data was captured in 50 cities during several months, daytimes, and good weather conditions. It was originally recorded as video so the frames were manually selected to have the following features: large number of dynamic objects, varying scene layout, and varying background.
	\item \textbf{CamVid}\cite{Brostow2008}\cite{Brostow2009}\footnote{\url{http://mi.eng.cam.ac.uk/research/projects/VideoRec/CamVid/}}: is a road/driving scene understanding database which was originally captured as five video sequences with a $960\times720$ resolution camera mounted on the dashboard of a car. Those sequences were sampled (four of them at 1 fps and one at 15 fps) adding up to $701$ frames. Those stills were manually annotated with $32$ classes: void, building, wall, tree, vegetation, fence, sidewalk, parking block, column/pole, traffic cone, bridge, sign, miscellaneous text, traffic light, sky, tunnel, archway, road, road shoulder, lane markings (driving), lane markings (non-driving), animal, pedestrian, child, cart luggage, bicyclist, motorcycle, car, SUV/pickup/truck, truck/bus, train, and other moving object. It is important to remark the partition introduced by Sturgess \emph{et al.}\cite{Sturgess2009} which divided the dataset into $367/100/233$ training, validation, and testing images respectively. That partition makes use of a subset of class labels: building, tree, sky, car, sign, road, pedestrian, fence, pole, sidewalk, and bicyclist.
	\item \textbf{KITTI}\cite{Geiger2013}: is one of the most popular datasets for use in mobile robotics and autonomous driving. It consists of hours of traffic scenarios recorded with a variety of sensor modalities, including high-resolution RGB, grayscale stereo cameras, and a 3D laser scanner. Despite its popularity, the dataset itself does not contain ground truth for semantic segmentation. However, various researchers have manually annotated parts of the dataset to fit their necessities. Álvarez \emph{et al.}\cite{Alvarez2012}\cite{Ros2015a} generated ground truth for $323$ images from the road detection challenge with three classes: road, vertical, and sky. Zhang \emph{et al.}\cite{Zhang2015} annotated $252$ ($140$ for training and $112$ for testing) acquisitions -- RGB and Velodyne scans -- from the tracking challenge for ten object categories: building, sky, road, vegetation, sidewalk, car, pedestrian, cyclist, sign/pole, and fence. Ros \emph{et al.} \cite{Ros2015} labeled $170$ training images and $46$ testing images (from the visual odometry challenge) with $11$ classes: building, tree, sky, car, sign, road, pedestrian, fence, pole, sidewalk, and bicyclist.
	\item \textbf{Youtube-Objects}\cite{Prest2012} is a database of videos collected from YouTube which contain objects from ten PASCAL \acs{VOC} classes: aeroplane, bird, boat, car, cat, cow, dog, horse, motorbike, and train. That database  does not contain pixel-wise annotations but Jain \emph{et al.}\cite{Jain2014} manually annotated a subset of $126$ sequences. They took every 10th frame from those sequences and generated semantic labels. That totals  $10167$ annotated frames at $480\times360$ pixels resolution.
	\item \textbf{Adobe's Portrait Segmentation}\cite{Shen2016}\footnote{\url{http://xiaoyongshen.me/webpage_portrait/index.html}}: this is a dataset of $800\times600$ pixels portrait images collected from Flickr, mainly captured with mobile front-facing cameras. The database consist of $1500$ training images and $300$ reserved for testing, both sets are fully binary annotated: person or background. The images were labeled in a semi-automatic way: first a face detector was run on each image to crop them to $600\times800$ pixels and then persons were manually annotated using Photoshop quick selection. This dataset is remarkable due to its specific purpose which makes it suitable for person in foreground segmentation applications.
	\item \textbf{\ac{MINC}}\cite{Bell2015}: this work is a dataset for patch material classification and full scene material segmentation. The dataset provides segment annotations for $23$ categories: wood, painted, fabric, glass, metal, tile, sky, foliage, polished stone, carpet, leather, mirror, brick, water, other, plastic, skin, stone, ceramic, hair, food, paper, and wallpaper. It contains $7061$ labeled material segmentations for training, $5000$ for test, and $2500$ for validation. The main source for these images is the OpenSurfaces dataset \cite{Bell2013}, which was augmented using other sources of imagery such as Flickr or Houzz. For that reason, image resolution for this dataset varies. On average, image resolution is approximately  $800\times500$ or $500\times800$.
	\item \textbf{\ac{DAVIS}}\cite{Perazzi2016}\cite{Pont-Tuset2017}\footnote{\url{http://davischallenge.org/index.html}}: this challenge is purposed for video object segmentation. Its dataset is composed by $50$ high-definition sequences which add up to $4219$ and $2023$ frames for training and validation respectively. Frame resolution varies across sequences but all of them were downsampled to $480$p for the challenge. Pixel-wise annotations are provided for each frame for four different categories: human, animal, vehicle, and object. Another feature from this dataset is the presence of at least one target foreground object in each sequence. In addition, it is designed not to have many different objects with significant motion. For those scenes which do have more than one target foreground object from the same class, they provide separated ground truth for each one of them to allow instance segmentation.
	\item \textbf{Stanford background}\cite{Gould2009}\footnote{\url{http://dags.stanford.edu/data/iccv09Data.tar.gz}}: dataset with outdoor scene images imported from existing public datasets: LabelMe, MSRC, PASCAL VOC and Geometric Context. The dataset contains 715 images (size of $320\times240$ pixels)  with at least one foreground object and having the horizon position within the image. The dataset is pixel-wise annotated (horizon location, pixel semantic class, pixel geometric class and image region) for evaluating methods for semantic scene understanding.
	\item \textbf{SiftFlow} \cite{Liu2009}: contains 2688 fully annotated images which are a subset of the LabelMe database \cite{Russell2008}. Most of the images are based on 8 different outdoor scenes including streets, mountains, fields, beaches and buildings. Images are $256\times256$ belonging to one of the 33 semantic classes. Unlabeled pixels, or pixels labeled as a different semantic class are treated as unlabeled. 
\end{itemize}

\subsection{\acs{2.5D} Datasets}

With the advent of low-cost range scanners, datasets including not only \acs{RGB} information but also depth maps are gaining popularity and usage. In this section, we review the most well-known \acs{2.5D} databases which include that kind of depth data.

\begin{itemize}
	\item \textbf{NYUDv2} \cite{Silberman2012}\footnote{\url{http://cs.nyu.edu/~silberman/projects/indoor_scene_seg_sup.html}}: this database consists of $1449$ indoor RGB-D images captured with a Microsoft Kinect device. It provides per-pixel dense labeling (category and instance levels) which were coalesced into 40 indoor object classes by Gupta \emph{et al.}\cite{Gupta2013} for both training ($795$ images) and testing ($654$) splits. This dataset is specially remarkable due to its indoor nature, this makes it really useful for certain robotic tasks at home. However, its relatively small scale with regard to other existing datasets hinders its application for deep learning architectures.	
	\item \textbf{SUN3D} 
	\cite{Xiao2013}\footnote{\url{http://sun3d.cs.princeton.edu/}}:
	similar to the NYUDv2, this dataset contains a large-scale \ac{RGB-D} video database, with 8 annotated sequences. Each frame has a semantic segmentation of the objects in the scene and information about the camera pose. It is still in progress and it will be composed by 415 sequences captured in 254 different spaces, in 41 different buildings. Moreover, some places have been captured multiple times at different moments of the day.
	\item \textbf{SUNRGBD}
	\cite{Song2015}\footnote{\url{http://rgbd.cs.princeton.edu/}}:
	captured with four \ac{RGB-D} sensors, this dataset contains 10000 RGB-D images, at a similar scale as PASCAL VOC. It contains images from NYU depth v2 \cite{Silberman2012}, Berkeley B3DO \cite{Janoch2013}, and SUN3D \cite{Xiao2013}. The whole dataset is densely annotated, including polygons, bounding boxes with orientation as well as a 3D room layout and category, being suitable for scene understanding tasks.
	\item \textbf{The Object Segmentation Database (OSD)}
	\cite{richtsfeld2012object}\footnote{\url{http://www.acin.tuwien.ac.at/?id=289}}
	this database has been designed for segmenting unknown objects from generic scenes even under partial occlusions. This dataset contains 111 entries, and provides depth image and color images together withper-pixel annotations for each one to evaluate object segmentation approaches. However, the dataset does not differentiate the category of different objects so its classes are reduced to a binary set of objects and not objects.

	\item \textbf{RGB-D Object Dataset}\cite{lai2011large}\footnote{\url{http://rgbd-dataset.cs.washington.edu/}}:
	this dataset is composed by video sequences of 300 common household objects organized in 51 categories arranged using WordNet hypernym-hyponym relationships. The dataset has been recorded using a Kinect style 3D camera that records synchronized and aligned $640\times480$ RGB and depth images at $30Hz$.	For each frame, the dataset provides, the RGB-D and depth images, a cropped ones containing the object, the location and a mask with per-pixel annotation. Moreover, each object has been placed on a turntable, providing isolated video sequences around 360 degrees. For the validation process, 22 annotated video sequences of natural indoor scenes containing the objects are provided.

\end{itemize}

\subsection{\acs{3D} Datasets}

Pure \acl{3D} databases are scarce, this kind of datasets usually provide \ac{CAD} meshes or other volumetric representations, such as point clouds. Generating large-scale \acs{3D} datasets for segmentation is costly and difficult, and not many deep learning methods are able to process that kind of data as it is. For those reasons, \acs{3D} datasets are not quite popular at the moment. In spite of that fact, we describe the most promising ones for the task at hand.

\begin{itemize}
	\item \textbf{ShapeNet Part}\cite{Yi2016}\footnote{\url{http://cs.stanford.edu/~ericyi/project_page/part_annotation/}}: is a subset of the ShapeNet\cite{Chang2015} repository which focuses on fine-grained 3D object segmentation. It contains $31,693$ meshes sampled from $16$ categories of the original dataset (airplane, earphone, cap, motorbike, bag, mug, laptop, table, guitar, knife, rocket, lamp, chair, pistol, car, and skateboard). Each shape class is labeled with two to five parts (totalling $50$ object parts across the whole dataset), e.g., each shape from the airplane class is labeled with wings, body, tail, and engine. Ground-truth labels are provided on points sampled from the meshes.

	\item \textbf{Stanford 2D-3D-S}\cite{Armeni2017}\footnote{\url{http://buildingparser.stanford.edu}}: is a multi-modal and large-scale indoor spaces dataset extending the Stanford 3D Semantic Parsing work \cite{Armeni2016}. It provides a variety of registered modalities -- 2D (RGB), 2.5D (depth maps and surface normals), and 3D (meshes and point clouds) --  with semantic annotations. The database is composed of $70,496$ full high-definition RGB images ($1080\times1080$ resolution) along with their corresponding depth maps, surface normals, meshes, and point clouds with semantic annotations (per-pixel and per-point). That data were captured in six indoor areas from three different educational and office buildings. That makes a total of $271$ rooms and approximately $700$ million points annotated with labels from $13$ categories: ceiling, floor, wall, column, beam, window, door, table, chair, bookcase, sofa, board, and clutter.

	\item \textbf{A Benchmark for 3D Mesh Segmentation}\cite{Chen:2009:ABF}\footnote{\url{http://segeval.cs.princeton.edu/}}: this benchmark is composed by 380 meshes classified in 19 categories (human, cup, glasses, airplane, ant, chair, octopus, table, teddy, hand, plier, fish, bird, armadillo, bust, mech, bearing, vase, fourleg). Each mesh has been manually segmented into functional parts, the main goal is to provide a sample distribution over "how humans decompose each mesh into functional parts".

	\item \textbf{Sydney Urban Objects Dataset}\cite{quadros2012feature}\footnote{\url{http://www.acfr.usyd.edu.au/papers/SydneyUrbanObjectsDataset.shtml}}: this dataset contains a variety of common urban road objects scanned with a Velodyne HDK-64E LIDAR. There are 631 individual scans (point clouds) of objects across classes of vehicles, pedestrians, signs and trees. The interesting point of this dataset is that, for each object, apart from the individual scan, a full 360-degrees annotated scan is provided.


	\item \textbf{Large-Scale Point Cloud Classification Benchmark} \cite{hackel2016contour}\footnote{\url{http://www.semantic3d.net/}}: this benchmark provides manually annotated 3D point clouds of diverse natural and urban scenes: churches, streets, railroad tracks, squares, villages, soccer fields, castles among others. This dataset features statically captured point clouds with very fine details and density. It contains $15$ large-scale point clouds for training and another $15$ for testing. Its scale can be grasped by the fact that it totals more than one billion labelled points.
\end{itemize}

\begin{table*}[!t]
	\centering
	\caption{Summary of semantic segmentation methods based on deep learning.}
	\label{table:methods}
	\resizebox{\linewidth}{!}{
	\begin{tabular}{|c|c|c|c|c|c|c|c|c|c|c|}
		\hline
		& & \multicolumn{7}{c|}{Targets} & &\\
		\hline
		\textbf{Name and Reference} & \textbf{Architecture} & \textbf{Accuracy} & \textbf{Efficiency} & \textbf{Training} & \textbf{Instance} & \textbf{Sequences} & \textbf{Multi-modal} & \textbf{3D} & \textbf{Source Code} & \textbf{Contribution(s)}\\
		\hline
		\acl{FCN}\cite{Long2015} & \acs{VGG}-16(\acs{FCN}) & $\star$ & $\star$ & $\star$ & \xmark & \xmark & \xmark & \xmark & \cmark & Forerunner\\
		SegNet\cite{Badrinarayanan2015} & \acs{VGG}-16 + Decoder & $\star\star\star$ & $\star\star$ & $\star$ & \xmark &\xmark & \xmark & \xmark & \cmark & Encoder-decoder\\
		Bayesian SegNet\cite{Kendall2015} & SegNet & $\star\star\star$ & $\star$ & $\star$ & \xmark & \xmark & \xmark & \xmark & \cmark & Uncertainty modeling\\
		DeepLab\cite{Chen2014a}\cite{Chen2016} & \acs{VGG}-16/ResNet-101 & $\star\star\star$ & $\star$ & $\star$ & \xmark & \xmark & \xmark & \xmark & \cmark & Standalone CRF, atrous convolutions\\
		\acs{MINC}-\acs{CNN} \cite{Bell2015} & GoogLeNet(\acs{FCN}) & $\star$ & $\star$ & $\star$ & \xmark & \xmark & \xmark & \xmark & \cmark & Patchwise \acs{CNN}, Standalone \acs{CRF}\\
		\acs{CRF}as\acs{RNN}\cite{Zheng2015} & \acs{FCN}-8s & $\star$ & $\star\star$ & $\star\star\star$ & \xmark & \xmark & \xmark & \xmark & \cmark & \acs{CRF} reformulated as \acs{RNN}\\
		Dilation\cite{Yu2015} & \acs{VGG}-16 & $\star\star\star$ & $\star$ & $\star$ & \xmark & \xmark & \xmark & \xmark & \cmark & Dilated convolutions\\
		ENet\cite{Paszke2016} & ENet bottleneck & $\star\star$ & $\star\star\star$ & $\star$ & \xmark & \xmark & \xmark & \xmark & \cmark & Bottleneck module for efficiency\\
		Multi-scale-CNN-Raj\cite{Raj2015} & \acs{VGG}-16(FCN) & $\star\star\star$ & $\star$ & $\star$ & \xmark & \xmark & \xmark & \xmark & \xmark & Multi-scale architecture\\
		Multi-scale-CNN-Eigen\cite{Eigen2015} & Custom & $\star\star\star$ & $\star$ & $\star$ & \xmark & \xmark & \xmark & \xmark & \cmark & Multi-scale sequential refinement\\
		Multi-scale-CNN-Roy\cite{Roy2016} & Multi-scale-CNN-Eigen & $\star\star\star$ & $\star$ & $\star$ & \xmark & \xmark & $\star\star$ & \xmark & \xmark & Multi-scale coarse-to-fine refinement\\
		Multi-scale-CNN-Bian\cite{Bian2016} & \acs{FCN} & $\star\star$ & $\star$ & $\star\star$ & \xmark & \xmark & \xmark & \xmark & \xmark & Independently trained multi-scale \acsp{FCN}\\
		ParseNet\cite{Liu2015} & \acs{VGG}-16 & $\star\star\star$ & $\star$ & $\star$ & \xmark & \xmark & \xmark & \xmark  & \cmark & Global context feature fusion\\ 
		ReSeg\cite{Visin2016} & \acs{VGG}-16 + ReNet & $\star\star$ & $\star$ & $\star$ & \xmark & \xmark & \xmark & \xmark & \cmark & Extension of ReNet to semantic segmentation\\
		\acs{LSTM-CF}\cite{Li2016b} & Fast R-CNN + DeepMask & $\star\star\star$ & $\star$ & $\star$ & \xmark & \xmark & \xmark & \xmark & \cmark & Fusion of contextual information from multiple sources\\
		2D-LSTM\cite{Byeon2015} & \acs{MDRNN} & $\star\star$ & $\star\star$ & $\star$ & \xmark & \xmark & \xmark & \xmark & \xmark & Image context modelling\\
		rCNN\cite{Pinheiro2014} & \acs{MDRNN} & $\star\star\star$ & $\star\star$ & $\star$ & \xmark & \xmark & \xmark & \xmark & \cmark & Different input sizes, image context\\
		DAG-RNN\cite{Shuai2015} & Elman network & $\star\star\star$ & $\star$ & $\star$ & \xmark & \xmark & \xmark & \xmark & \cmark & Graph image structure for context modelling\\
		\acs{SDS}\cite{Hariharan2014} & \acs{R-CNN} + Box \acs{CNN} & $\star\star\star$ & $\star$ & $\star$ & $\star\star$ & \xmark & \xmark & \xmark  & \cmark & Simultaneous detection and segmentation\\
		DeepMask\cite{Pinheiro2015} & \acs{VGG}-A & $\star\star\star$ & $\star$ & $\star$ & $\star\star$ & \xmark & \xmark & \xmark & \cmark & Proposals generation for segmentation\\
		SharpMask\cite{Pinheiro2016} & DeepMask & $\star\star\star$ & $\star$ & $\star$ & $\star\star\star$ & \xmark & \xmark & \xmark & \cmark & Top-down refinement module\\
		MultiPathNet\cite{Zagoruyko2016} & Fast R-CNN + DeepMask & $\star\star\star$ & $\star$ & $\star$ & $\star\star\star$ & \xmark & \xmark & \xmark & \cmark & Multi path information flow through network\\
		Huang-3DCNN\cite{Huang2016} & Own \acs{3D}\acs{CNN} & $\star$ & $\star$ & $\star$ & \xmark & \xmark & \xmark & $\star\star\star$ & \xmark & \acs{3D}\acs{CNN} for voxelized point clouds\\
		PointNet\cite{Qi2016} & Own \acs{MLP}-based & $\star\star$ & $\star$ & $\star$ & \xmark & \xmark & \xmark & $\star\star\star$ & \cmark & Segmentation of unordered point sets\\
		Clockwork Convnet\cite{Shelhamer2016} & \acs{FCN} & $\star\star$ & $\star\star$ & $\star$ & \xmark & $\star\star\star$ &\xmark & \xmark & \cmark & Clockwork scheduling for sequences\\
		\acs{3D}\acs{CNN}-Zhang & Own \acs{3D}\acs{CNN} & $\star\star$ & $\star$ & $\star$ & \xmark & $\star\star\star$ & \xmark & \xmark & \cmark & \acs{3D} convolutions and graph cut for sequences\\
		End2End Vox2Vox\cite{Tran2016} & \acs{C3D} & $\star\star$ & $\star$ & $\star$ & \xmark & $\star\star\star$ & \xmark & \xmark & \xmark & 3D convolutions/deconvolutions for sequences\\
		\hline
	\end{tabular}}
\end{table*}
\section{Methods}
\label{sec:methods}

The relentless success of deep learning techniques in various high-level computer vision tasks -- in particular, supervised approaches such as \acfp{CNN} for image classification or object detection \cite{Krizhevsky2012}\cite{Simonyan2014}\cite{Szegedy2015} --  motivated researchers to explore the capabilities of such networks for pixel-level labelling problems like semantic segmentation. The key advantage of these deep learning techniques, which gives them an edge over traditional methods, is the ability to learn appropriate feature representations for the problem at hand, e.g., pixel labelling on a particular dataset, in an end-to-end fashion instead of using hand-crafted features that require domain expertise, effort, and often too much fine-tuning to make them work on a particular scenario.

\begin{figure}[!hbt]
	\centering
	\includegraphics[width=\linewidth]{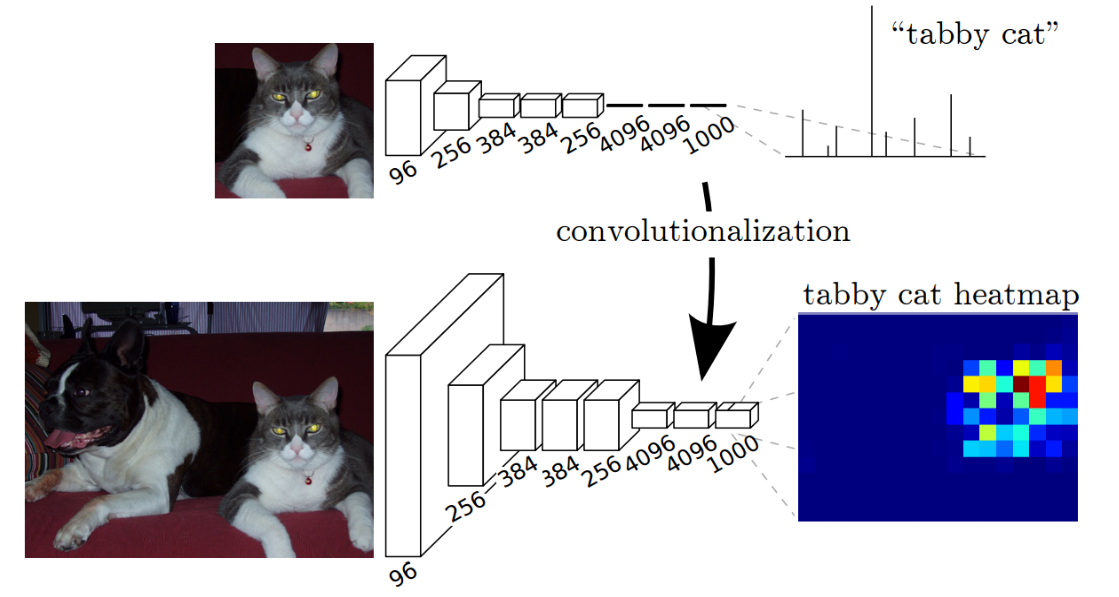}
	\includegraphics[width=0.86\linewidth]{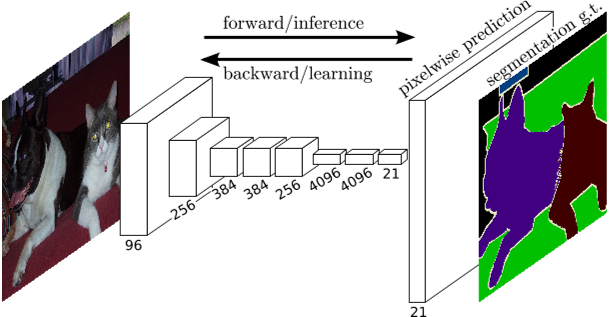}
	\caption{\acl{FCN} figure by Long \emph{et al.}\cite{Long2015}. Transforming a classification-purposed \ac{CNN} to produce spatial heatmaps by replacing fully connected layers with convolutional ones. Including a deconvolution layer for upsampling allows dense inference and learning for per-pixel labeling.}
	\label{fig:convolutionalization}
\end{figure}
\begin{figure*}[!t]
	\centering
	\includegraphics[width=0.8\linewidth]{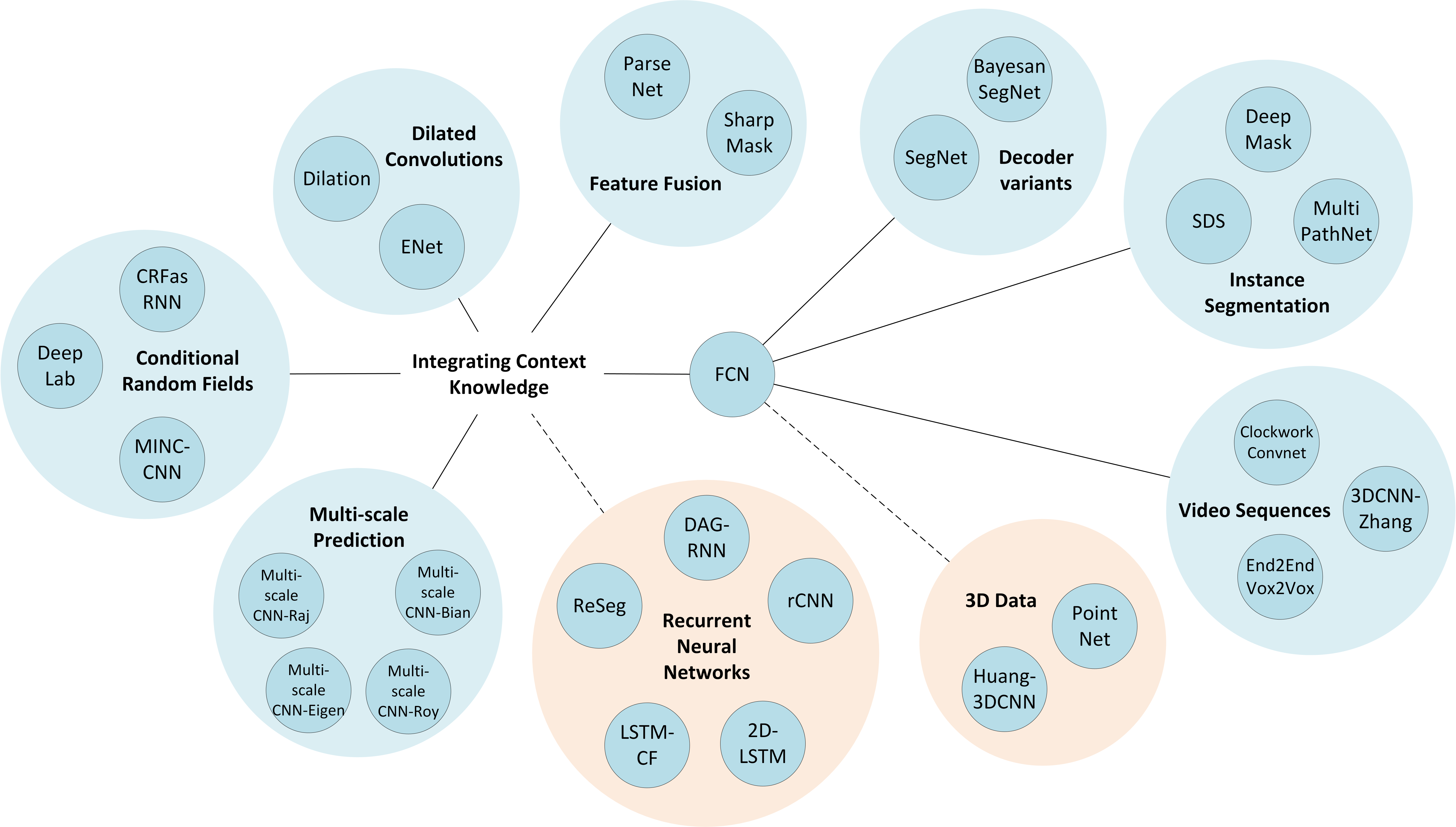}
	\caption{Visualization of the reviewed methods.}
	\label{fig:graph}
\end{figure*}

Currently, the most successful state-of-the-art deep learning techniques for semantic segmentation stem from a common forerunner: the \emph{\acf{FCN}} by Long \emph{et al.}\cite{Long2015}. The insight of that approach was to take advantage of existing \acp{CNN} as powerful visual models that are able to learn hierarchies of features. They transformed those existing and well-known classification models -- AlexNet\cite{Krizhevsky2012}, \ac{VGG} (16-layer net)\cite{Simonyan2014}, GoogLeNet\cite{Szegedy2015}, and ResNet \cite{He2016} -- into fully convolutional ones by replacing the fully connected layers with convolutional ones to output spatial maps instead of classification scores. Those maps are upsampled using fractionally strided convolutions (also named deconvolutions \cite{Zeiler2011}\cite{Zeiler2014}) to produce dense per-pixel labeled outputs. This work is considered a milestone since it showed how \acp{CNN} can be trained end-to-end for this problem, efficiently learning how to make dense predictions for semantic segmentation with inputs of arbitrary sizes. This approach achieved a significant improvement in segmentation accuracy over traditional methods on standard datasets like PASCAL \ac{VOC}, while preserving efficiency at inference. For all those reasons, and other significant contributions, the \ac{FCN} is the cornerstone of deep learning applied to semantic segmentation. The convolutionalization process is shown in Figure \ref{fig:convolutionalization}.

Despite the power and flexibility of the \acs{FCN} model, it still lacks various features which hinder its application to certain problems and situations: its inherent spatial invariance does not take into account useful global context information, no instance-awareness is present by default, efficiency is still far from real-time execution at high resolutions, and it is not completely suited for unstructured data such as \acs{3D} point clouds or models. Those problems will be reviewed in this section, as well as the state-of-the-art solutions that have been proposed in the literature to overcome those hurdles. Table \ref{table:methods} provides a summary of that review. It shows all reviewed methods (sorted by appearance order in the section), their base architecture, their main contribution, and a classification depending on the target of the work: accuracy, efficiency, training simplicity, sequence processing, multi-modal inputs, and \acs{3D} data. Each target is graded from one to three stars ($\star$) depending on how much focus puts the work on it, and a mark (\xmark) if that issue is not addressed. In addition, Figure \ref{fig:graph} shows a graph of the reviewed methods for the sake of visualization.

\subsection{Decoder Variants}

Apart from the \acs{FCN} architecture, other variants were developed to transform a network whose purpose was classification to make it suitable for segmentation. Arguably, \acs{FCN}-based architectures are more popular and successful, but other alternatives are also remarkable. In general terms, all of them take a network for classification, such as \acs{VGG}-16, and remove its fully connected layers. This part of the new segmentation network often receives the name of encoder and produce low-resolution image representations or feature maps. The problem lies on learning to decode or map those low-resolution images to pixel-wise predictions for segmentation. This part is named decoder and it is usually the divergence point in this kind of architectures.

\begin{figure}[!hbt]
	\includegraphics[width=\linewidth]{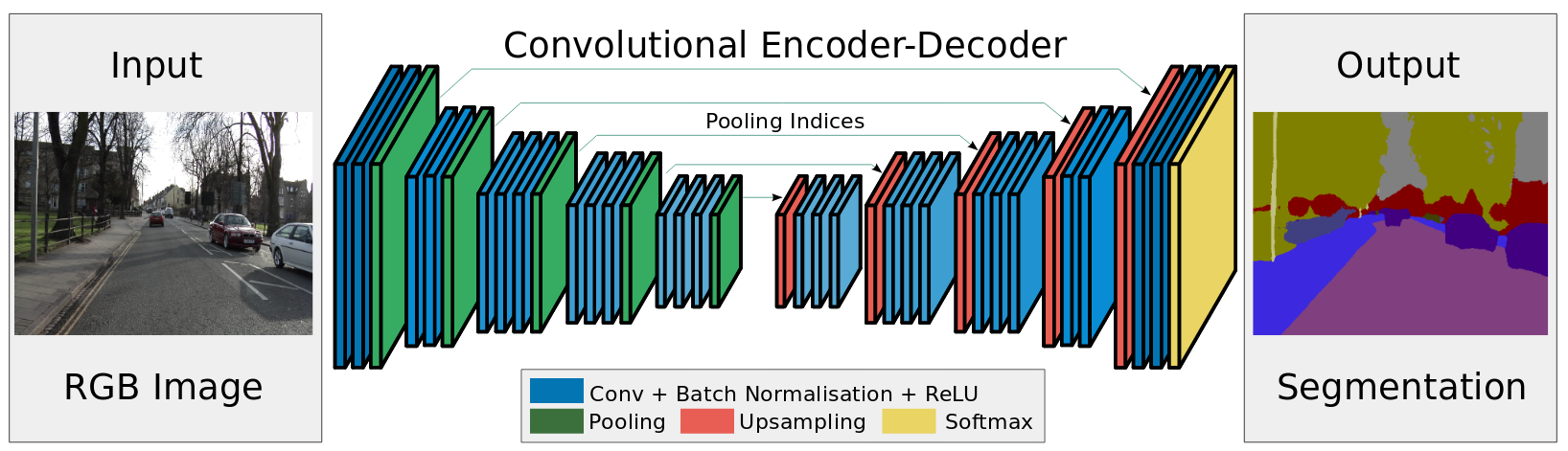}
	\caption{SegNet architecture with an encoder and a decoder followed by a softmax classifier for pixel-wise classification. Figure extracted from \cite{Badrinarayanan2015}.}
	\label{fig:segnet}
\end{figure}

SegNet\cite{Badrinarayanan2015} is a clear example of this divergence (see Figure \ref{fig:segnet}). The decoder stage of SegNet is composed by a set of upsampling and convolution layers which are at last followed by a softmax classifier to predict pixel-wise labels for an output which has the same resolution as the input image. Each upsampling layer in the decoder stage corresponds to a max-pooling one in the encoder part. Those layers upsample feature maps using the max-pooling indices from their corresponding feature maps in the encoder phase. The upsampled maps are then convolved with a set of trainable filter banks to produce dense feature maps. When the feature maps have been restored to the original resolution, they are fed to the softmax classifier to produce the final segmentation.

\begin{figure}[!hbt]
	\includegraphics[width=\linewidth]{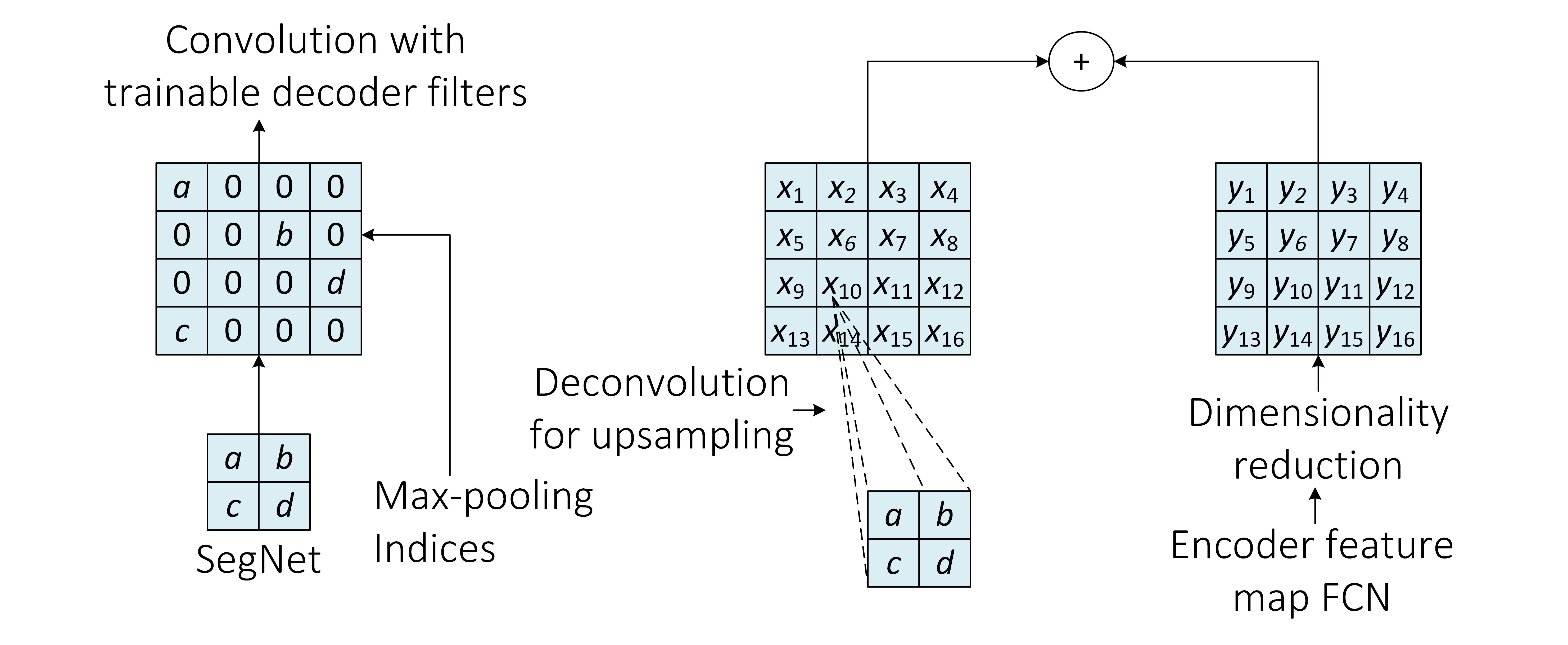}
	\caption{Comparison of SegNet (left) and \acs{FCN} (right) decoders. While SegNet uses max-pooling indices from the corresponding encoder stage to upsample, \acs{FCN} learns deconvolution filters to upsample (adding the corresponding feature map from the encoder stage). Figure reproduced from \cite{Badrinarayanan2015}.}
	\label{fig:segnetvsfcn}
\end{figure}

On the other hand, \acs{FCN}-based architectures make use of learnable deconvolution filters to upsample feature maps. After that, the upsampled feature maps are added element-wise to the corresponding feature map generated by the convolution layer in the encoder part. Figure \ref{fig:segnetvsfcn} shows a comparison of both approaches. 

\subsection{Integrating Context Knowledge}

Semantic segmentation is a problem that requires the integration of information from various spatial scales. It also implies balancing local and global information. On the one hand, fine-grained or local information is crucial to achieve good pixel-level accuracy. On the other hand, it is also important to integrate information from the global context of the image to be able to resolve local ambiguities.

Vanilla \acp{CNN} struggle with this balance. Pooling layers, which allow the networks to achieve some degree of spatial invariance and keep computational cost at bay, dispose of the global context information. Even purely \acp{CNN} -- without pooling layers -- are limited since the receptive field of their units can only grow linearly with the number of layers.

Many approaches can be taken to make \acp{CNN} aware of that global information: refinement as a post-processing step with \acp{CRF}, dilated convolutions, multi-scale aggregation, or even defer the context modeling to another kind of deep networks such as \acp{RNN}.

\subsubsection{\aclp{CRF}}

As we mentioned before, the inherent invariance to spatial transformations of \ac{CNN} architectures limits the very same spatial accuracy for segmentation tasks. One possible and common approach to refine the output of a segmentation system and boost its ability to capture fine-grained details is to apply a post-processing stage using a \acf{CRF}. \acp{CRF} enable the combination of low-level image information -- such as the interactions between pixels \cite{Rother2004}\cite{Shotton2009} -- with the output of multi-class inference systems that produce per-pixel class scores. That combination is especially important to capture long-range dependencies, which \acp{CNN} fail to consider, and fine local details.

\begin{figure}[!t]
	\hfill
	\begin{subfigure}{0.19\linewidth}
		\includegraphics[width=\linewidth]{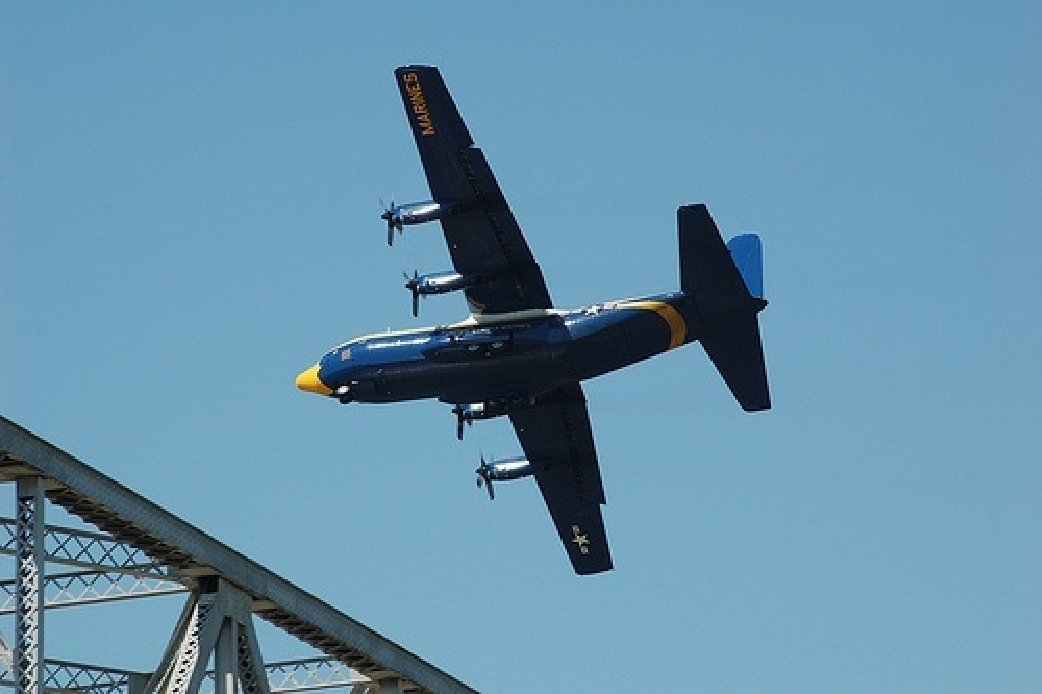}\\
		\includegraphics[width=\linewidth]{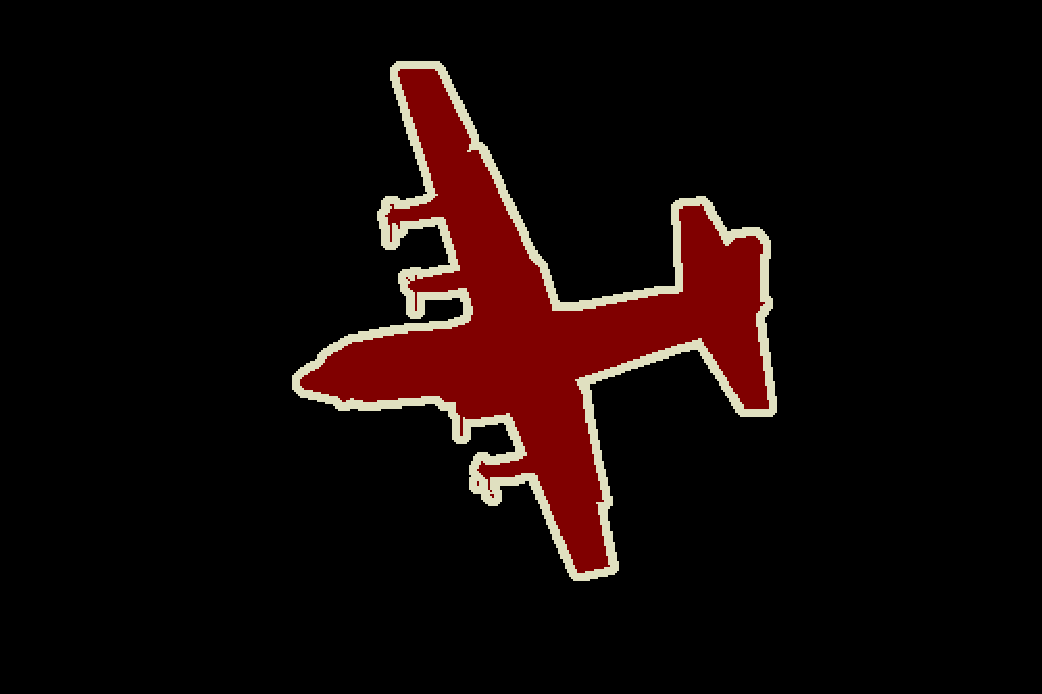}
		\caption{GT}
	\end{subfigure}
	\hfill
	\begin{subfigure}{0.19\linewidth}
		\includegraphics[width=\linewidth]{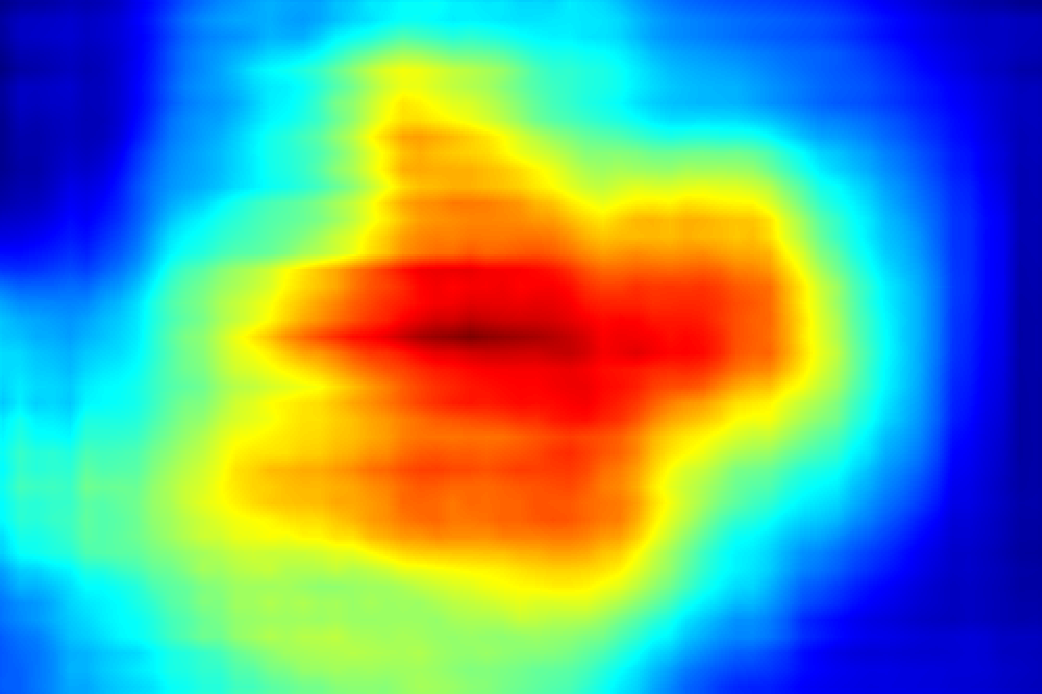}\\
		\includegraphics[width=\linewidth]{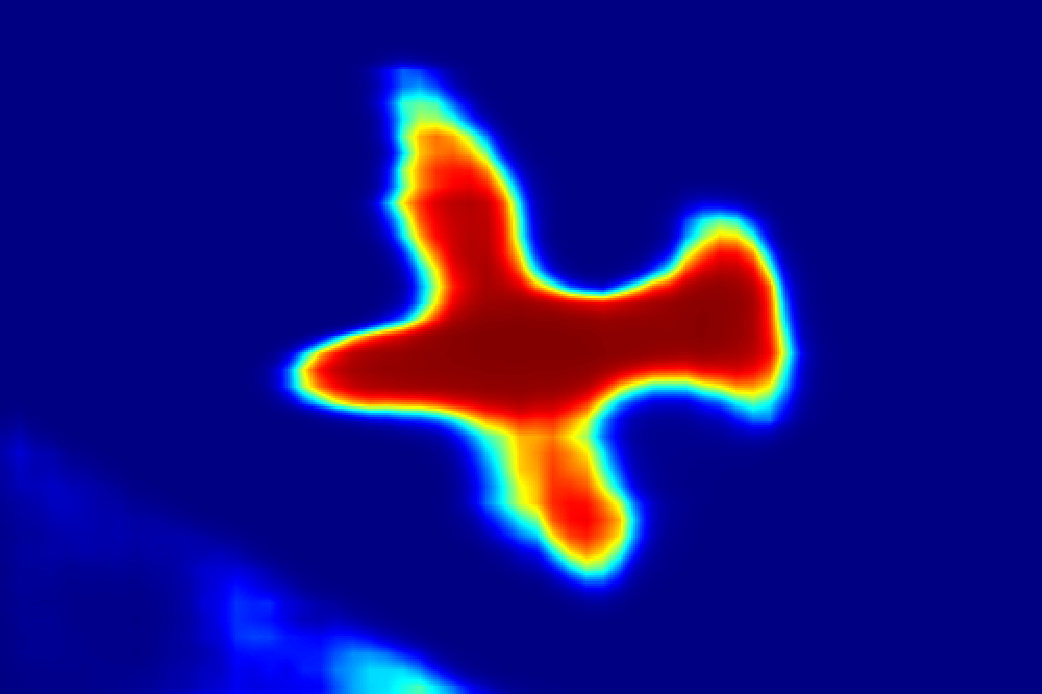}
		\caption{\ac{CNN}out}
	\end{subfigure}
	\hfill
	\begin{subfigure}{0.19\linewidth}
		\includegraphics[width=\linewidth]{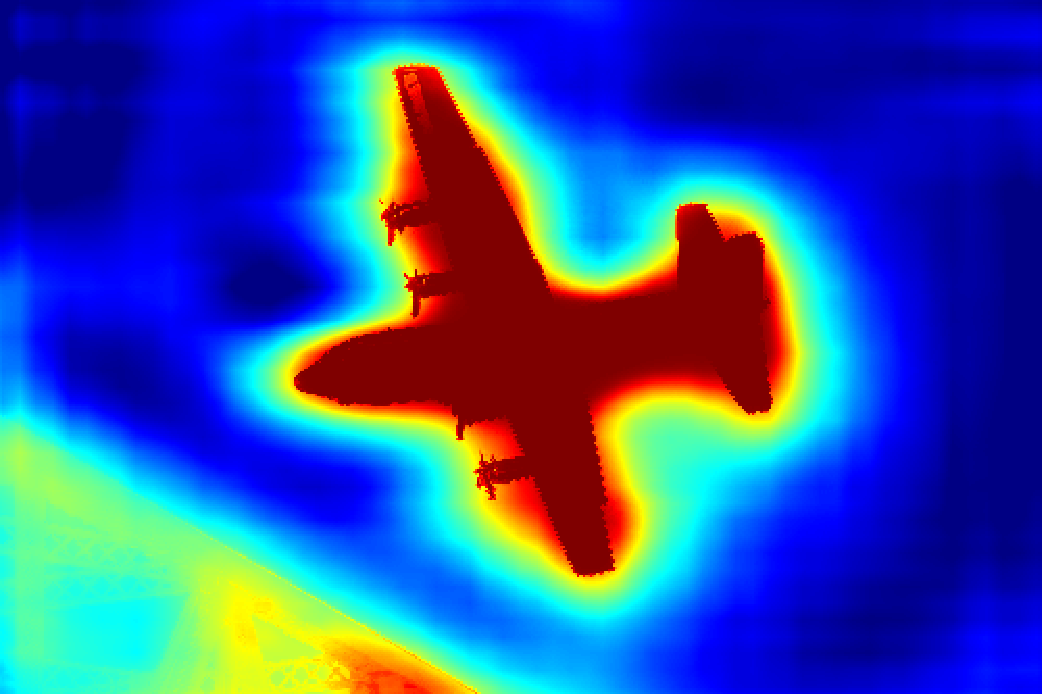}\\
		\includegraphics[width=\linewidth]{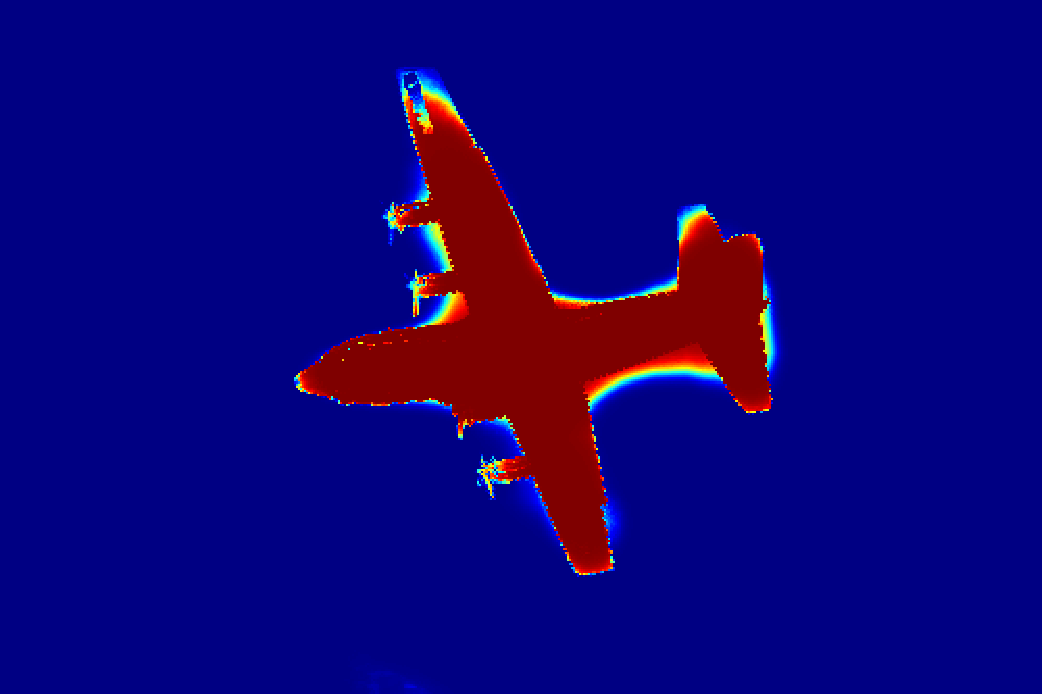}
		\caption{\ac{CRF}it1}
	\end{subfigure}		
	\hfill
	\begin{subfigure}{0.19\linewidth}
		\includegraphics[width=\linewidth]{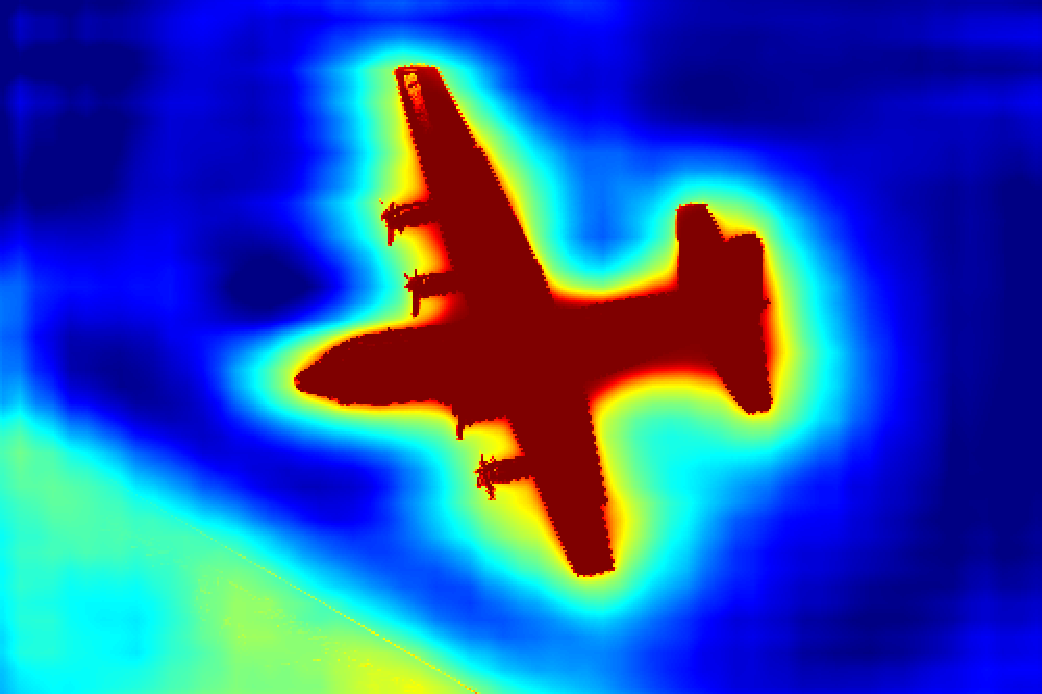}\\
		\includegraphics[width=\linewidth]{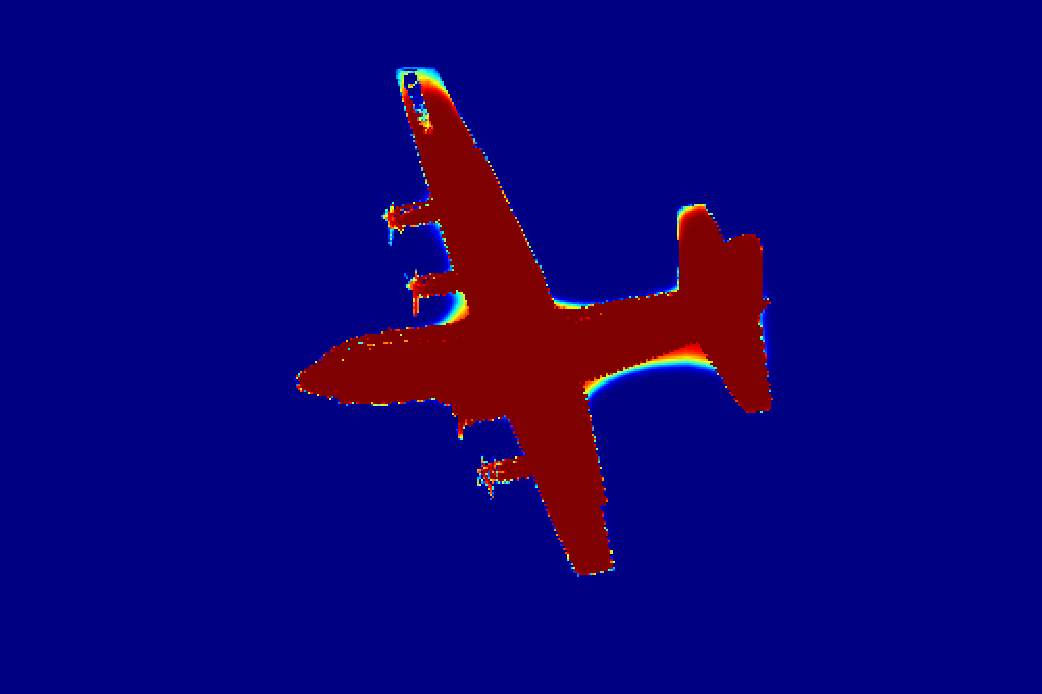}
		\caption{\ac{CRF}it2}
	\end{subfigure}		
	\hfill
	\begin{subfigure}{0.19\linewidth}
		\includegraphics[width=\linewidth]{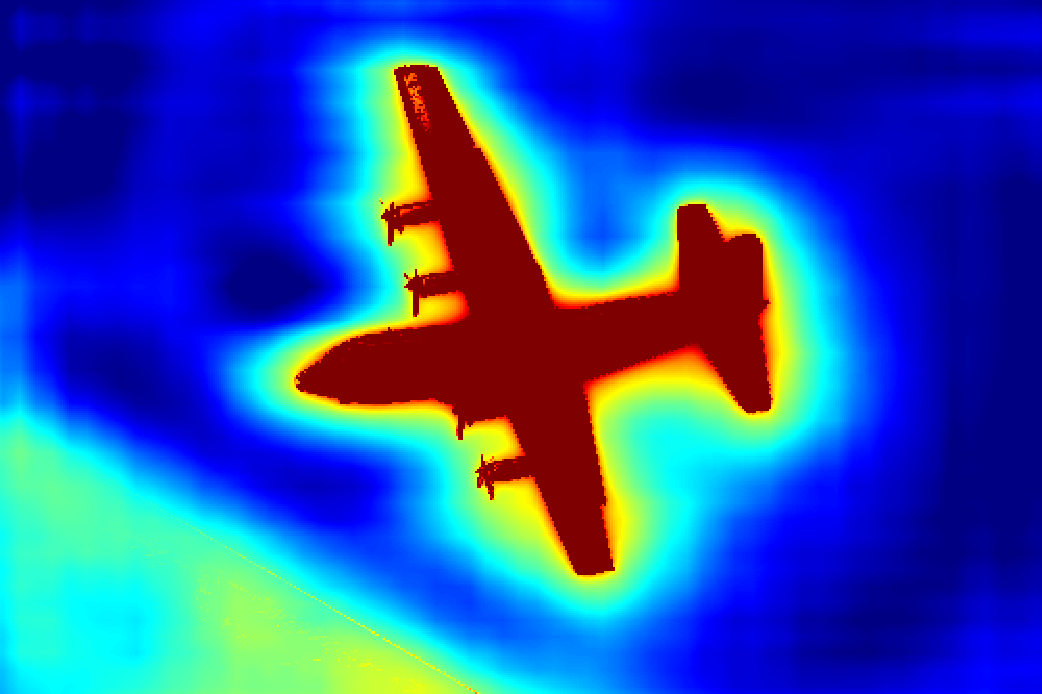}\\
		\includegraphics[width=\linewidth]{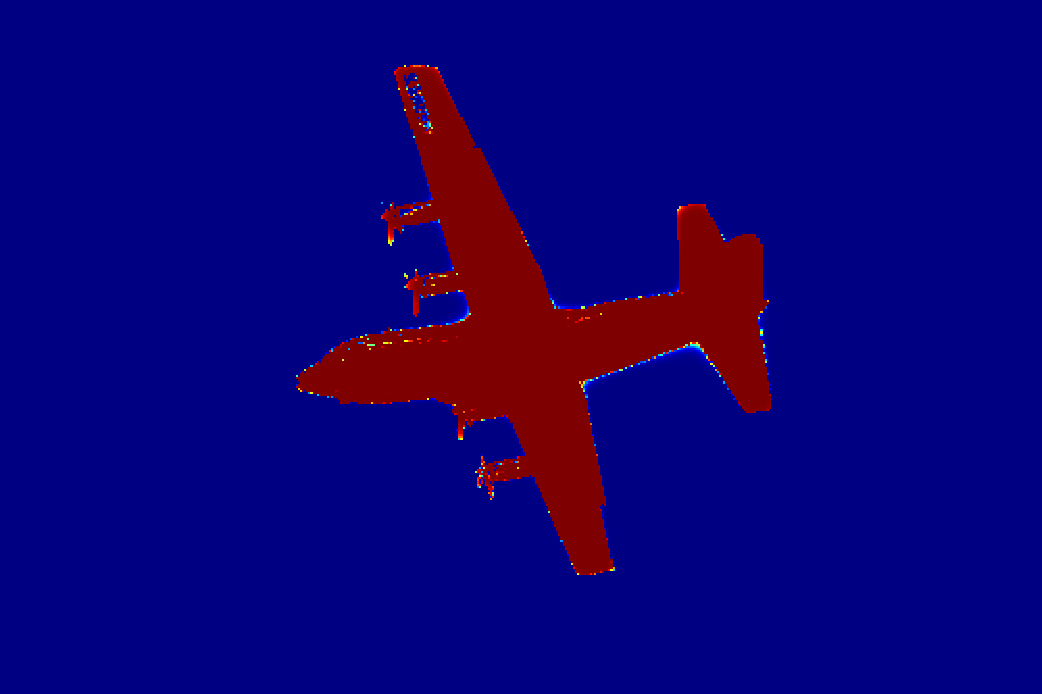}
		\caption{\ac{CRF}it10}
	\end{subfigure}	
	\hfill
	\caption{\ac{CRF} refinement per iteration as shown by the authors of DeepLab\cite{Chen2014a}. The first row shows the score maps (inputs before the softmax function) and the second one shows the belief maps (output of the softmax function).}
	\label{fig:crf-deeplab}
\end{figure}

The DeepLab models \cite{Chen2014a}\cite{Chen2016} make use of the fully connected pairwise \ac{CRF} by Krähenbühl and Koltun\cite{Koltun2011}\cite{Kraehenbuehl2013} as a separated post-processing step in their pipeline to refine the segmentation result. It models each pixel as a node in the field and employs one pairwise term for each pair of pixels no matter how far they lie (this model is known as dense or fully connected factor graph). By using this model, both short and long-range interactions are taken into account, rendering the system able to recover detailed structures in the segmentation that were lost due to the spatial invariance of the \ac{CNN}. Despite the fact that usually fully connected models are inefficient, this model can be efficiently approximated via probabilistic inference. Figure \ref{fig:crf-deeplab} shows the effect of this \ac{CRF}-based post-processing on the score and belief maps produced by the DeepLab model.

The material recognition in the wild network by Bell \emph{et al.}\cite{Bell2015} makes use of various \acp{CNN} trained to identify patches in the \ac{MINC} database. Those \acp{CNN} are used on a sliding window fashion to classify those patches. Their weights are transferred to the same networks converted into \acp{FCN} by adding the corresponding upsampling layers. The outputs are averaged to generate a probability map. At last, the same \ac{CRF} from DeepLab, but discretely optimized, is applied to predict and refine the material at every pixel.

Another significant work applying a \ac{CRF} to refine the segmentation of a \ac{FCN} is the CRFasRNN by Zheng \emph{et al.}\cite{Zheng2015}. The main contribution of that work is the reformulation of the dense \ac{CRF} with pairwise potentials as an integral part of the network. By unrolling the mean-field inference steps as \acp{RNN}, they make it possible to fully integrate the \ac{CRF} with a \ac{FCN} and train the whole network end-to-end. This work demonstrates the reformulation of \acp{CRF} as \acp{RNN} to form a part of a deep network, in contrast with Pinheiro et al. \cite{Pinheiro2014} which employed \acp{RNN} to model large spatial dependencies.

\subsubsection{Dilated Convolutions}

\begin{figure}[!hbt]
	\hfill
	\begin{subfigure}{0.30\linewidth}
		\includegraphics[width=\linewidth]{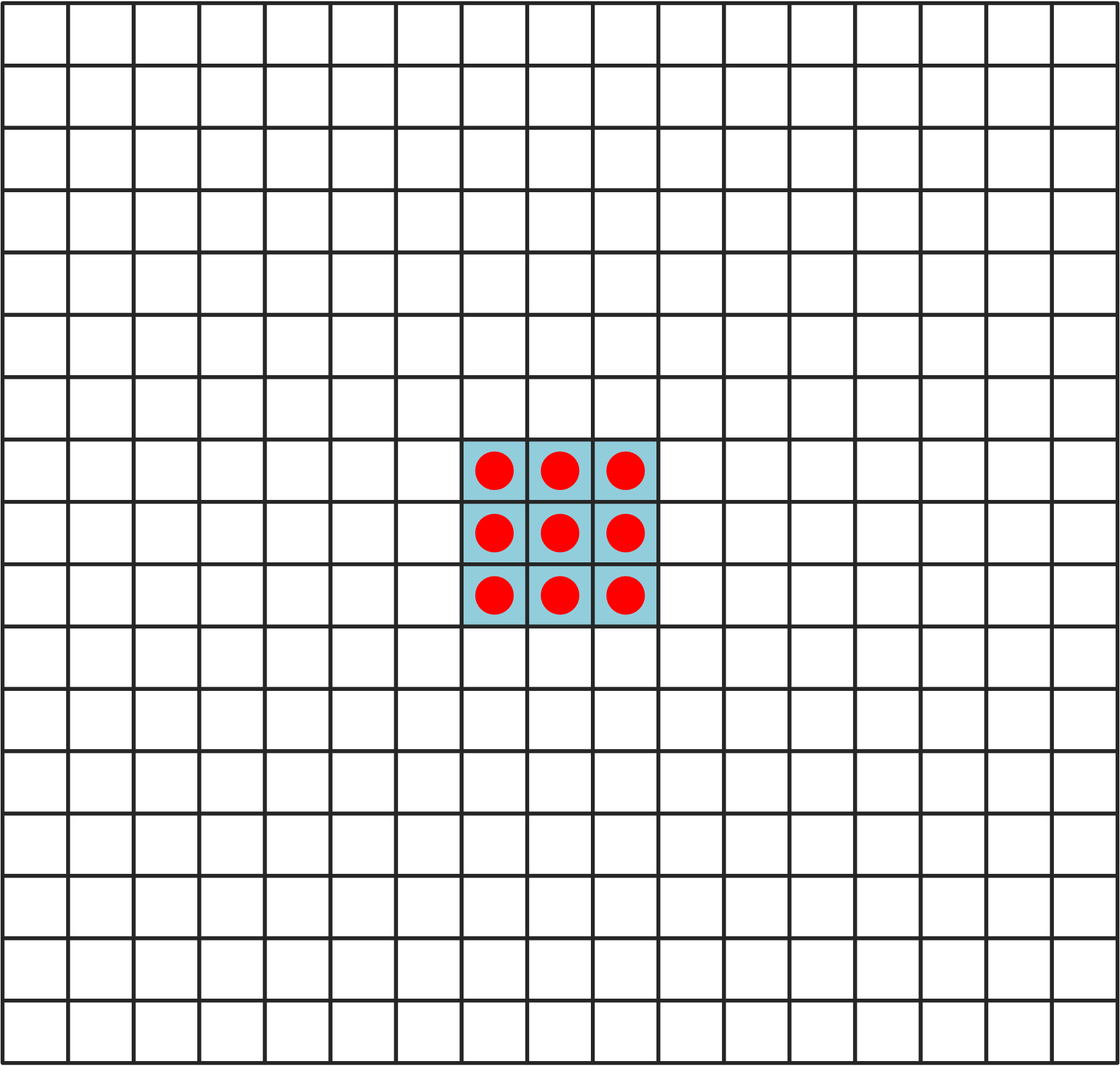}
		\caption{$1$-dilated}
		\label{fig:dilated-convolution:1}
	\end{subfigure}
	\hfill
	\begin{subfigure}{0.30\linewidth}
		\includegraphics[width=\linewidth]{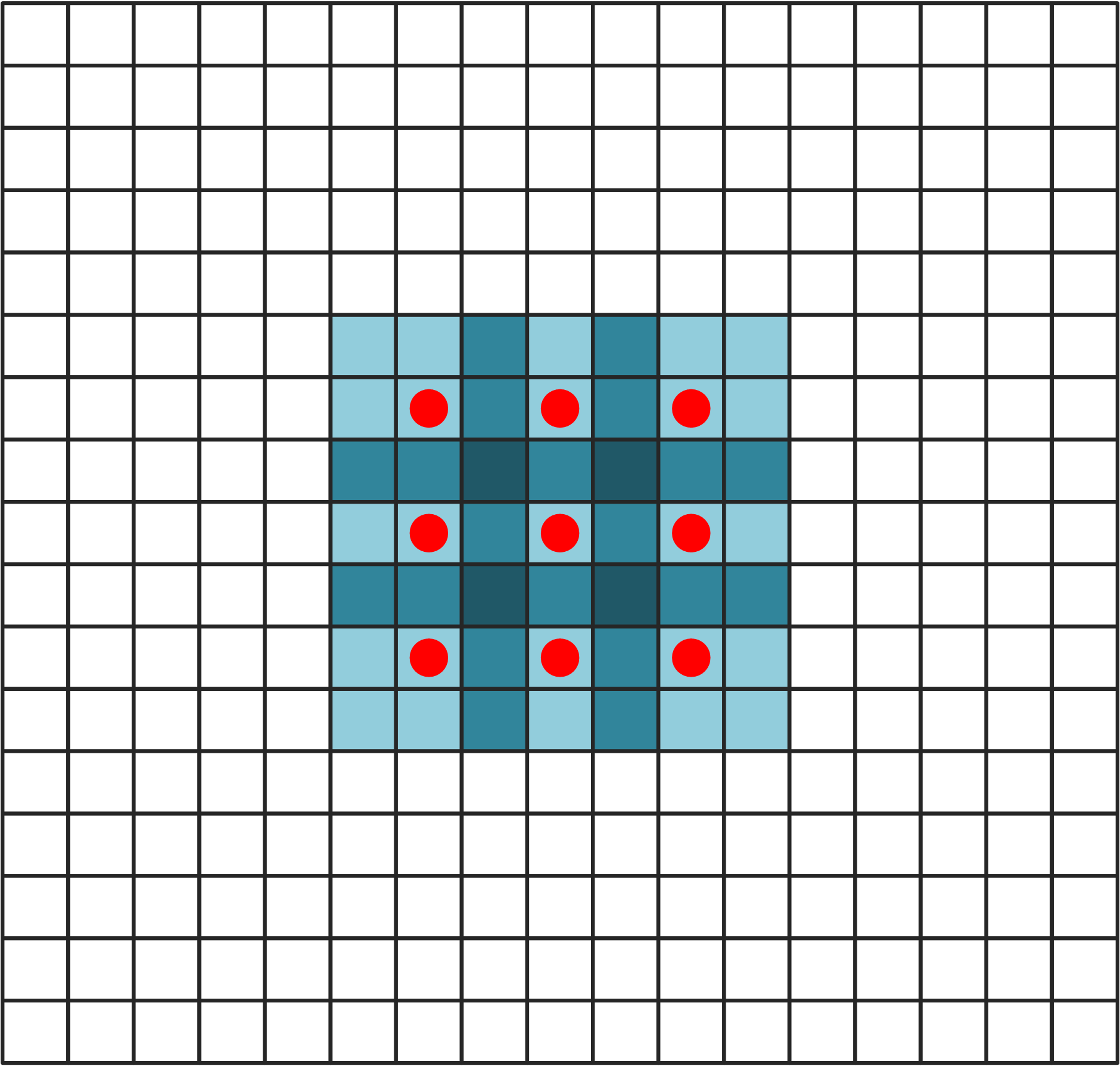}
		\caption{$2$-dilated}
		\label{fig:dilated-convolution:2}
	\end{subfigure}
	\hfill
	\begin{subfigure}{0.30\linewidth}
		\includegraphics[width=\linewidth]{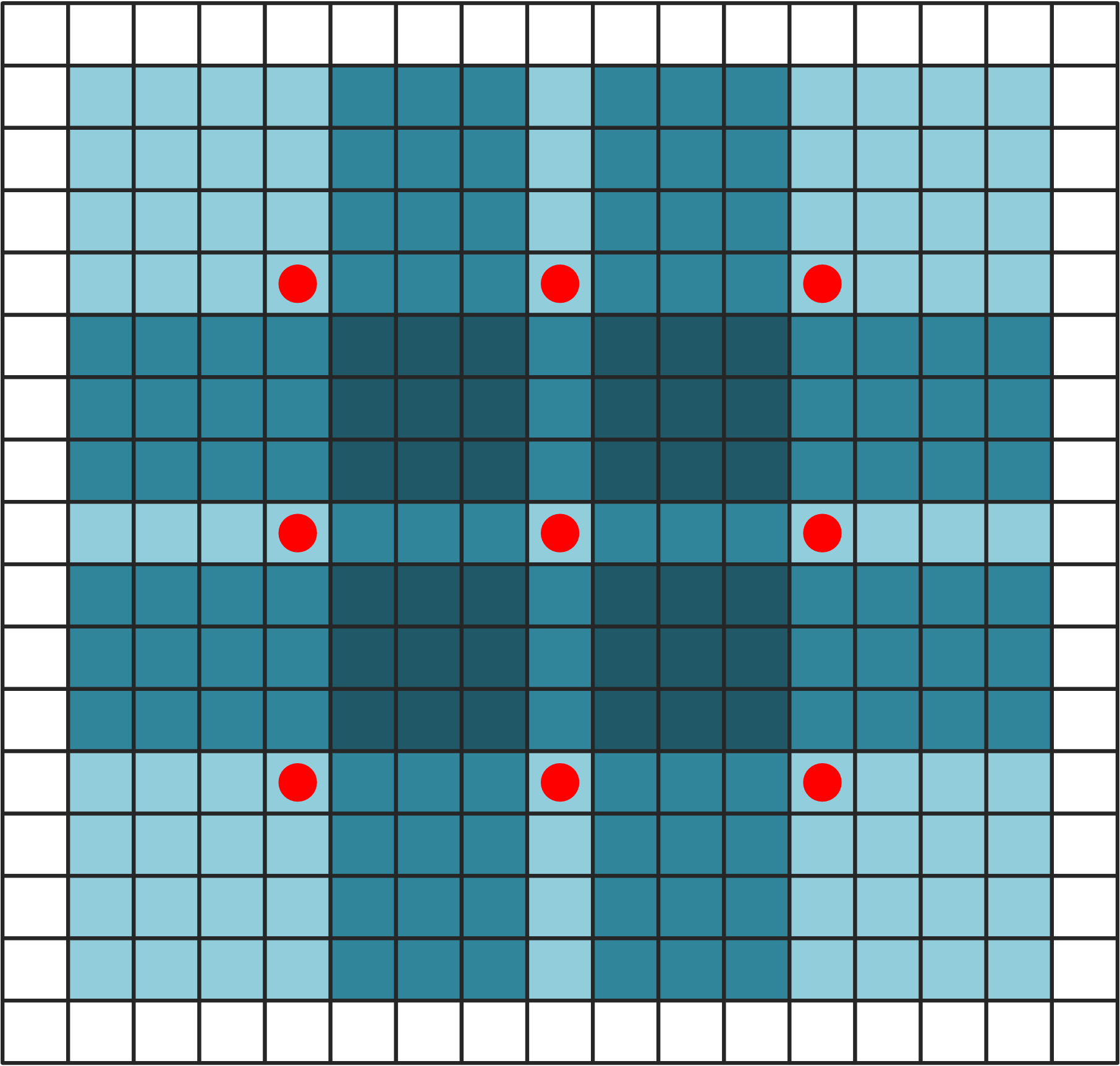}
		\caption{$3$-dilated}
		\label{fig:dilated-convolution:3}
	\end{subfigure}
	\hfill
	\caption{As shown in \cite{Yu2015}, dilated convolution filters with various dilation rates: (\protect\subref{fig:dilated-convolution:1}) $1$-dilated convolutions in which each unit has a $3\times3$ receptive fields, (\protect\subref{fig:dilated-convolution:2}) $2$-dilated ones with $7\times7$ receptive fields, and (\protect\subref{fig:dilated-convolution:3}) $3$-dilated convolutions with $15\times15$ receptive fields.}
	\label{fig:dilated-convolution}
\end{figure}

Dilated convolutions, also named \emph{à-trous} convolutions, are a generalization of Kronecker-factored convolutional filters \cite{Zhou2015} which support exponentially expanding receptive fields without losing resolution. In other words, dilated convolutions are regular ones that make use of upsampled filters. The dilation rate $l$ controls that upsampling factor. As shown in Figure \ref{fig:dilated-convolution}, stacking $l$-dilated convolution makes the receptive fields grow exponentially while the number of parameters for the filters keeps a linear growth. This means that dilated convolutions allow efficient dense feature extraction on any arbitrary resolution. As a side note, it is important to remark that typical convolutions are just $1$-dilated convolutions.

\begin{figure}[!hbt]
	\centering
	\includegraphics[width=\linewidth]{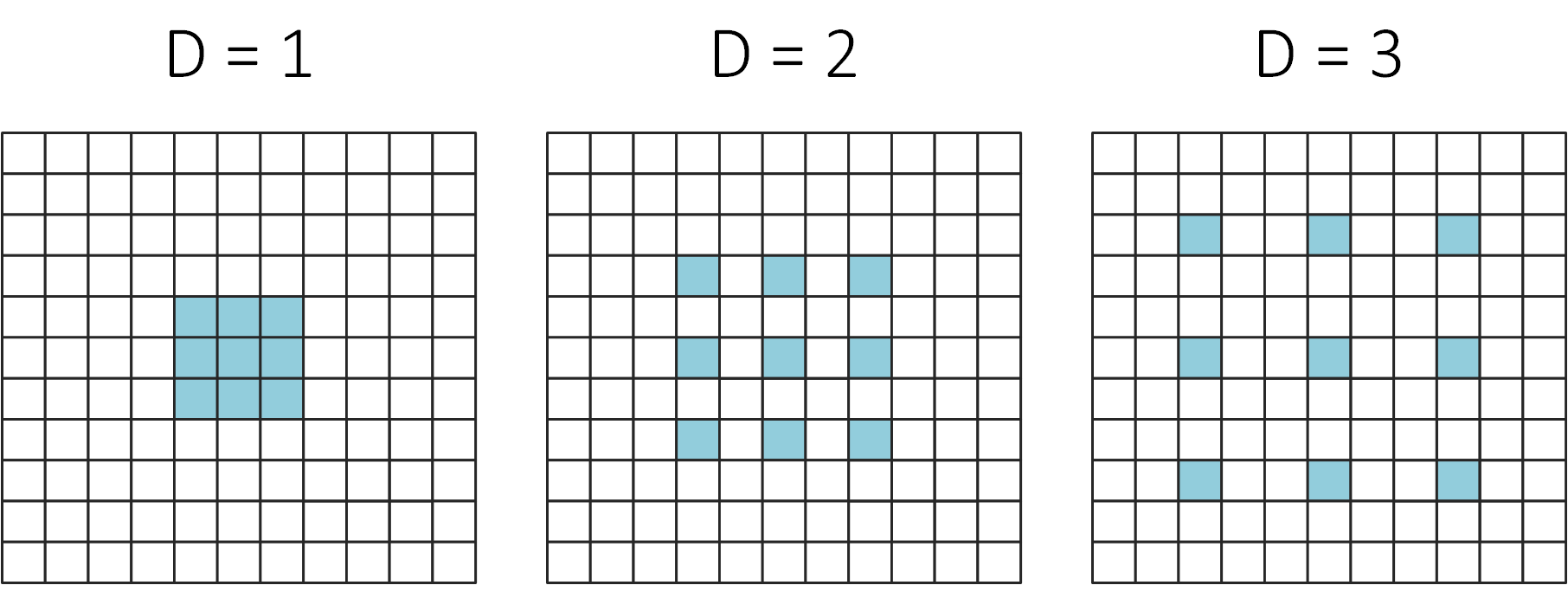}
	\caption{Filter elements (green) matched to input elements when using $3\times3$ dilated convolutions with various dilation rates. From left to right: $1, 2, $ and $3$.}
	\label{fig:dilated-convolution-filter}
\end{figure}

In practice, it is equivalent to dilating the filter before doing the usual convolution. That means expanding its size, according to the dilation rate, while filling the empty elements with zeros. In other words, the filter weights are matched to distant elements which are not adjacent if the dilation rate is greater than one. Figure \ref{fig:dilated-convolution-filter} shows examples of dilated filters.

The most important works that make use of dilated convolutions are the multi-scale context aggregation module by Yu \emph{et al.}\cite{Yu2015}, the already mentioned DeepLab (its improved version)\cite{Chen2016}, and the real-time network ENet\cite{Paszke2016}. All of them use combinations of dilated convolutions with increasing dilation rates to have wider receptive fields with no additional cost and without overly downsampling the feature maps. Those works also show a common trend: dilated convolutions are tightly coupled to multi-scale context aggregation as we will explain in the following section.

\subsubsection{Multi-scale Prediction}

Another possible way to deal with context knowledge integration is the use of multi-scale predictions. Almost every single parameter of a \acs{CNN} affects the scale of the generated feature maps. In other words, the very same architecture will have an impact on the number of pixels of the input image which correspond to a pixel of the feature map. This means that the filters will implicitly learn to detect features at specific scales (presumably with certain invariance degree). Furthermore, those parameters are usually tightly coupled to the problem at hand, making it difficult for the models to generalize to different scales. One possible way to overcome that obstacle is to use multi-scale networks which generally make use of multiple networks that target different scales and then merge the predictions to produce a single output.

Raj \emph{et al.}\cite{Raj2015} propose a multi-scale version of a fully convolutional \acs{VGG}-16. That network has two paths, one that processes the input at the original resolution and another one which doubles it. The first path goes through a shallow convolutional network. The second one goes through the fully convolutional \acs{VGG}-16 and an extra convolutional layer. The result of that second path is upsampled and combined with the result of the first path. That concatenated output then goes through another set of convolutional layers to generate the final output. As a result, the network becomes more robust to scale variations.

\begin{figure}[!htb]
	\centering
	\includegraphics[width=0.86\linewidth]{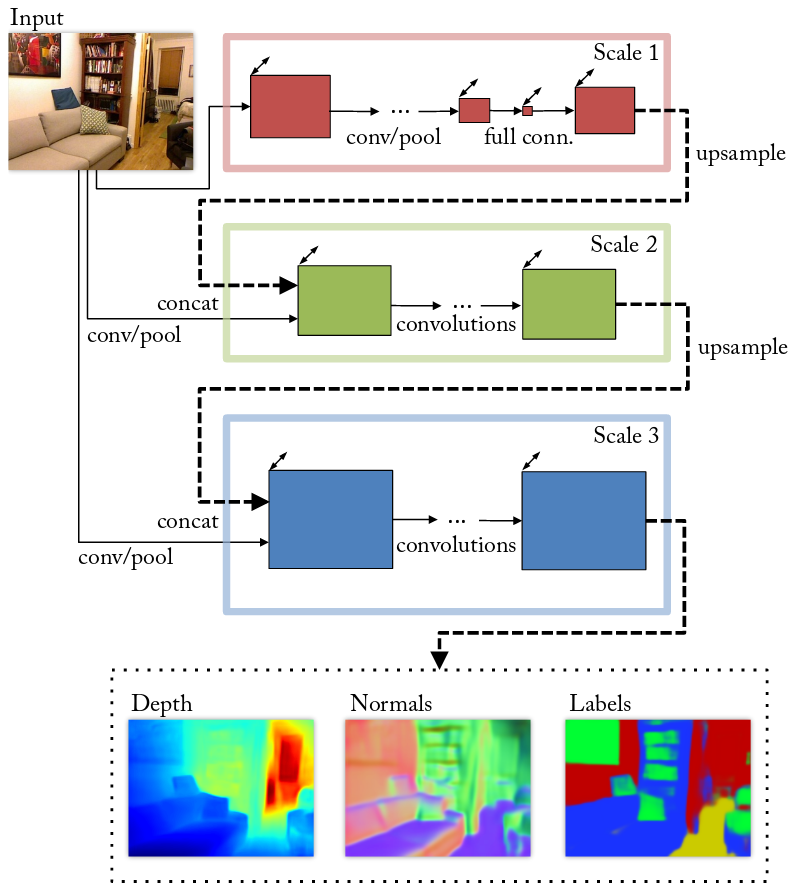}
	\caption{Multi-scale \acs{CNN} architecture proposed by Eigen \emph{et al.}\cite{Eigen2015}. The network progressively refines the output using a sequence of scales to estimate depth, normals, and also perform semantic segmentation over an \acs{RGB} input. Figure extracted from \cite{Eigen2015}.}
	\label{fig:eigen-multiscale}
\end{figure}

Roy \emph{et al.}\cite{Roy2016} take a different approach using a network composed by four multi-scale \acp{CNN}. Those four networks have the same architecture introduced by Eigen \emph{et al.} \cite{Eigen2015}. One of those networks is devoted to finding semantic labels for the scene. That network extracts features from a progressively coarse-to-fine sequence of scales (see Figure \ref{fig:eigen-multiscale}).

Another remarkable work is the network proposed by Bian \emph{et al.}\cite{Bian2016}. That network is a composition of $n$ \acp{FCN} which operate at different scales. The features extracted from the networks are fused together (after the necessary upsampling with an appropriate padding) and then they go through an additional convolutional layer to produce the final segmentation. The main contribution of this architecture is the two-stage learning process which involves, first, training each network independently, then the networks are combined and the last layer is fine-tuned. This multi-scale model allows to add an arbitrary number of newly trained networks in an efficient manner.

\subsubsection{Feature Fusion}

\begin{figure}[!hbt]
	\centering
	\includegraphics[width=0.4\linewidth]{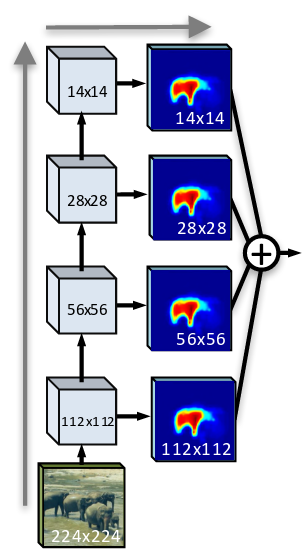}
	\caption{Skip-connection-like architecture, which performs late fusion of feature maps as if making independent predictions for each layer and merging the results. Figure extracted from \cite{Pinheiro2016}.}
	\label{fig:skipconnections}
\end{figure}

Another way of adding context information to a fully convolutional architecture for segmentation is feature fusion. This technique consists of merging a global feature (extracted from a previous layer in a network) with a more local feature map extracted from a subsequent layer. Common architectures such as the original \acs{FCN} make use of skip connections to perform a late fusion by combining the feature maps extracted from different layers (see Figure \ref{fig:skipconnections}).

\begin{figure}[!hbt]
	\centering
	\includegraphics[width=\linewidth]{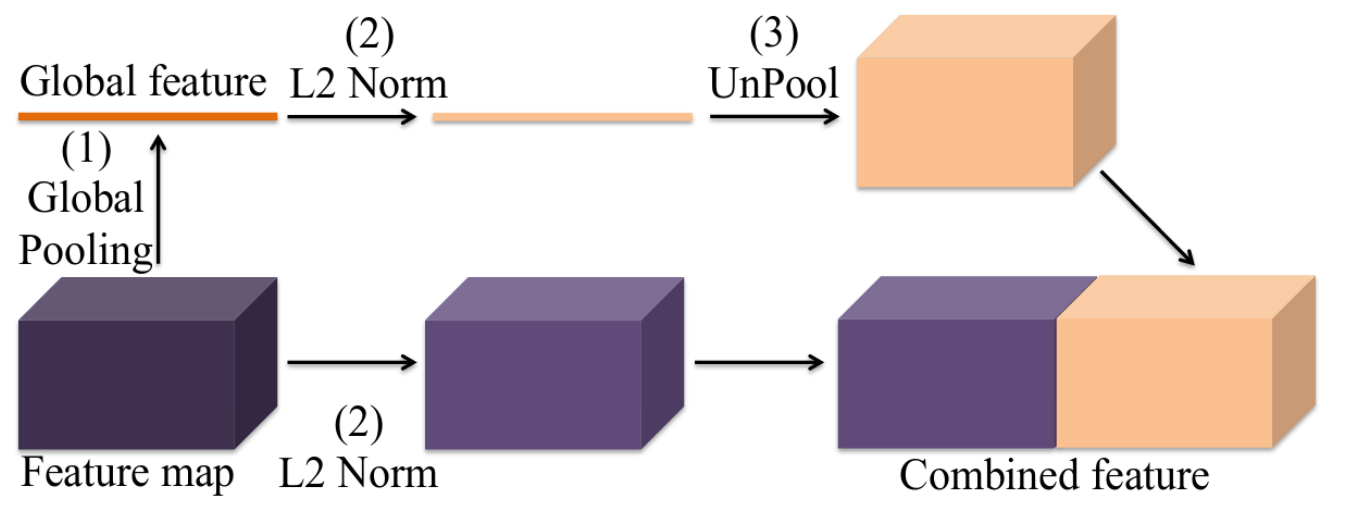}
	\caption{ParseNet context module overview in which a global feature (from a previous layer) is combined with the feature of the next layer to add context information. Figure extracted from \cite{Liu2015}.}
	\label{fig:parsenet-module}
\end{figure}

Another approach is performing early fusion. This approach is taken by ParseNet\cite{Liu2015} in their context module. The global feature is unpooled to the same spatial size as the local feature and then they are concatenated to generate a combined feature that is used in the next layer or to learn a classifier. Figure \ref{fig:parsenet-module} shows a representation of that process.

This feature fusion idea was continued by Pinheiro \emph{et al.} in their SharpMask network \cite{Pinheiro2016}, which introduced a progressive refinement module to incorporate features from the previous layer to the next in a top-down architecture. This work will be reviewed later since it is mainly focused on instance segmentation.

\subsubsection{\aclp{RNN}}

\begin{figure*}[!t]
	\centering
	\includegraphics[width=\linewidth]{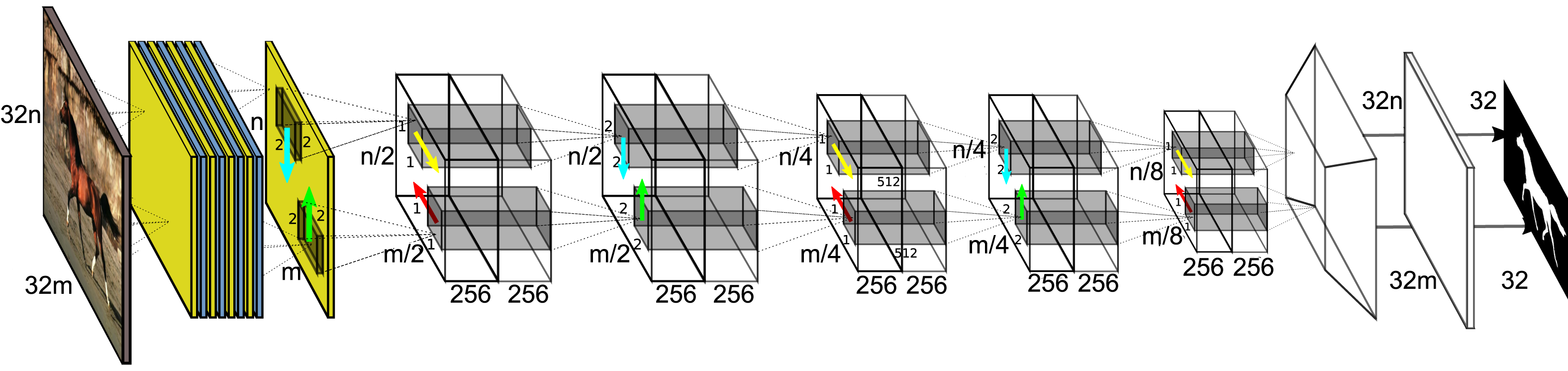}
	\caption{Representation of ReSeg network. VGG-16 convolutional layers are represented by the blue and yellow first layers. The rest of the architecture is based on the ReNet approach with fine-tuning purposes. Figure extracted from \cite{Visin2016}.}
	\label{fig:reseg-network}
\end{figure*}

As we noticed, \acp{CNN} have been successfully applied to multi-dimensional data, such as images. Nevertheless, these networks rely on hand specified kernels limiting the architecture to local contexts. Taking advantage of its topological structure, \aclp{RNN} have been successfully applied for modeling short- and long-temporal sequences. In this way and by linking together pixel-level and local information, \acp{RNN} are able to successfully model global contexts and improve semantic segmentation. However, one important issue is the lack of a natural sequential structure in images and the focus of standard vanilla \acp{RNN} architectures on one-dimensional inputs.  
	
Based on ReNet model for image classification Visin \emph{et al.}\cite{Visin2015} proposed an architecture for semantic segmentation called ReSeg \cite{Visin2016} represented in Figure \ref{fig:reseg-network}. In this approach, the input image is processed with the first layers of the \acs{VGG}-16 network \cite{Simonyan2014}, feeding the resulting feature maps into one or more ReNet layers for fine-tuning. Finally, feature maps are resized using upsampling layers based on transposed convolutions. In this approach \acp{GRU} have been used as they strike a good performance balance regarding memory usage and computational power. Vanilla \acsp{RNN} have problems modeling long-term dependencies mainly due to the vanishing gradients problem. Several derived models such as \ac{LSTM} networks \cite{Hochreiter1997} and \acp{GRU} \cite{Cho2014} are the state-of-art in this field to avoid such problem.

Inspired on the same ReNet architecture, a novel \ac{LSTM-CF} model for scene labeling was proposed by \cite{ZhenLi2016}. In this approach, they use two different data sources: \acs{RGB} and depth. The \acs{RGB} pipeline relies on a variant of the DeepLab architecture \cite{Chen2014} concatenating features at three different scales to enrich feature representation (inspired by \cite{Li2016}). The global context is modeled vertically over both, depth and photometric data sources, concluding with a horizontal fusion in both direction over these vertical contexts.

As we noticed, modeling image global contexts is related to 2D recurrent approaches by unfolding vertically and horizontally the network over the input images. Based on the same idea, Byeon et al. \cite{Byeon2015} purposed a simple 2D \ac{LSTM}-based architecture in which the input image is divided into non-overlapping windows which are fed into four separate \acp{LSTM} memory blocks. This work emphasizes its low computational complexity on a single-core CPU and the model simplicity. 

Another approach for capturing global information relies on using bigger input windows in order to model larger contexts. Nevertheless, this reduces images resolution and also implies several problems regarding to window overlapping. However, Pinheiro et al. \cite{Pinheiro2014} introduced \acp{rCNN} which recurrently train with different input window sizes taking into account previous predictions by using a different input window sizes. In this way, predicted labels are automatically smoothed increasing the performance.

Undirected cyclic graphs (UCGs) were also adopted to model image contexts for semantic segmentation \cite{Shuai2015}. Nevertheless, \acp{RNN} are not directly applicable to UCG and the solution is decomposing it into several directed graphs (DAGs). In this approach, images are processed by three different layers: image feature map produced by \acs{CNN}, model image contextual dependencies with DAG-RNNs, and deconvolution layer for upsampling feature maps. This work demonstrates how \acp{RNN} can be used together with graphs to successfully model long-range contextual dependencies, overcoming state-of-the-art approaches in terms of performance.

\subsection{Instance Segmentation}

Instance segmentation is considered the next step after semantic segmentation and at the same time the most challenging problem in comparison with the rest of low-level pixel segmentation techniques. Its main purpose is to represent objects of the same class splitted into different instances. The automation of this process is not straightforward, thus the number of instances is initially unknown and the evaluation of performed predictions is not pixel-wise such as in semantic segmentation. Consequently, this problem remains partially unsolved but the interest in this field is motivated by its potential applicability. Instance labeling provides us extra information for reasoning about occlusion situations, also counting the number of elements belonging to the same class and for detecting a particular object for grasping in robotics tasks, among many other applications.

For this purpose, Hariharan et al. \cite{Hariharan2014} proposed a \ac{SDS} method in order to improve performance over already existing works. Their pipeline uses, firstly, a bottom-up hierarchical image segmentation and object candidate generation process called \ac{MCG} \cite{Arbelaez2014} to obtain region proposals. For each region, features are extracted by using an adapted version of the \ac{R-CNN} \cite{Girshick2014}, which is fine-tuned using bounding boxes provided by the \ac{MCG} method instead of selective search and also alongside region foreground features. Then, each region proposal is classified by using a linear \ac{SVM} on top of the \acs{CNN} features. Finally, and for refinement purposes, \ac{NMS} is applied to the previous proposals.

Later, Pinheiro et al. \cite{Pinheiro2015} presented DeepMask model, an object proposal approach based on a single ConvNet. This model predicts a segmentation mask for an input patch and the likelihood of this patch for containing an object. The two tasks are learned jointly and computed by a single network, sharing most of the layers except last ones which are task-specific. 

\begin{figure}[!hbt]
	\centering
	\includegraphics[width=0.75\linewidth]{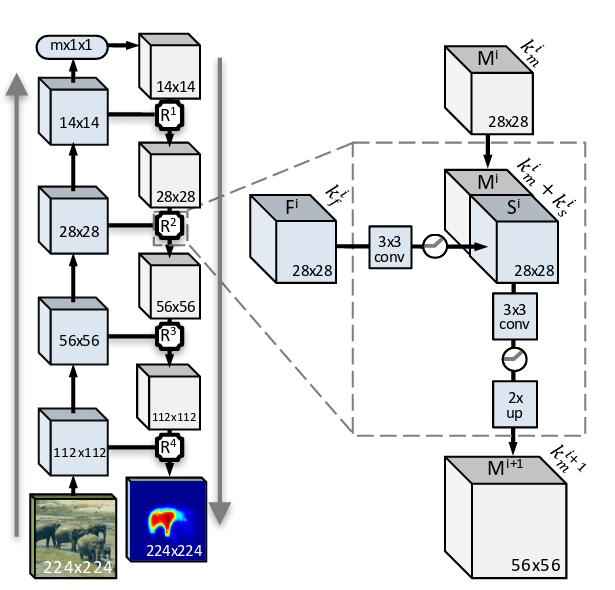}
	\caption{SharpMask's top-down architecture with progressive refinement using their signature modules. That refinement merges spatially rich information from lower-level features with high-level semantic cues encoded in upper layers. Figure extracted from \cite{Pinheiro2015}.}
	\label{fig:sharpmask-refinement-module}
\end{figure}

Based on the DeepMask architecture as a starting point due to its effectiveness, the same authors presented a novel architecture for object instance segmentation implementing a top-down refinement process \cite{Pinheiro2016} and achieving a better performance in terms of accuracy and speed. The goal of this process is to efficiently merge low-level features with high-level semantic information from upper network layers. The process consisted in different refinement modules stacked together (one module per pooling layer), with the purpose of inverting pooling effect by generating a new upsampled object encoding. Figure \ref{fig:sharpmask-refinement-module} shows the refinement module in SharpMask.

Another approach, based on Fast R-CNN as a starting point and using DeepMask object proposals instead of Selective Search was presented by Zagoruyko et al \cite{Zagoruyko2016}. This combined system called MultiPath classifier, improved performance over COCO dataset and supposed three modifications to Fast R-CNN: improving localization with an integral loss, provide context by using foveal regions and finally skip connections to give multi-scale features to the network. The system achieved a 66\% improvement over the baseline Fast R-CNN.

As we have seen, most of the methods mentioned above rely on existing object detectors limiting in this way model performance. Even so, instance segmentation process remains an unresolved research problem and the mentioned works are only a small part of this challenging research topic.

\subsection{\acs{RGB-D} Data}

As we noticed, a significant amount of work has been done in semantic segmentation by using photometric data. Nevertheless, the use of structural information was spurred on with the advent of low-cost \acs{RGB-D} sensors which provide useful geometric cues extracted from depth information. Several works focused on \acs{RGB-D} scene segmentation have reported an improvement in the fine-grained labeling precision by using depth information and not only photometric data. Using depth information for segmentation is considered more challenging because of the unpredictable variation of scene illumination alongside incomplete representation of objects due to complex occlusions. However, various works have successfully made use of depth information to increase accuracy.

The use of depth images with approaches focused on photometric data is not straightforward. Depth data needs to be encoded with three channels at each pixel as if it was an \acs{RGB} images. Different techniques such as \ac{HHA} \cite{Gupta2014} are used for encoding the depth into three channels as follows: horizontal disparity, height above ground, and the angle between local surface normal and the inferred gravity direction. In this way, we can input depth images to models designed for \acs{RGB} data and improve in this way the performance by learning new features from structural information. Several works such as \cite{ZhenLi2016} are based on this encoding technique.

In the literature, related to methods that use \acs{RGB-D} data, we can also find some works that leverage a multi-view approach to improve existing single-view works.

Zeng \emph{et al.}\cite{Zeng2016} present an object segmentation approach that leverages multi-view \acs{RGB-D} data and deep learning techniques. RGB-D images captured from each viewpoint are fed to a \acs{FCN} network which returns a 40-class probability for each pixel in each image. Segmentation labels are threshold by using three times the standard deviation above the mean probability across all views. Moreover, in this work, multiple networks for feature extraction were trained (AlexNet \cite{Krizhevsky2012} and \acs{VGG}-16 \cite{Simonyan2014}), evaluating the benefits of using depth information. They found that adding depth did not yield any major improvements in segmentation performance, which could be caused by noise in the depth information. The described approach was presented during the 2016 Amazon Picking Challenge. This work is a minor contribution towards multi-view deep learning systems since \acs{RGB} images are independently fed to a \acs{FCN} network.

Ma \emph{et al.}\cite{Ma2017} propose a novel approach for object-class segmentation using a multi-view deep learning technique. Multiple views are obtained from a moving RGB-D camera. During the training stage, camera trajectory is obtained using an RGB-D SLAM technique, then RGB-D images are warped into ground-truth annotated frames in order to enforce multi-view consistency for training. The proposed approach is based on FuseNet\cite{Hazirbas2016}, which combines \acs{RGB} and depth images for semantic segmentation, and improves the original work by adding multi-scale loss minimization.

\subsection{\acs{3D} Data}

\begin{figure}[!htb]
	\centering
	\includegraphics[width=\linewidth]{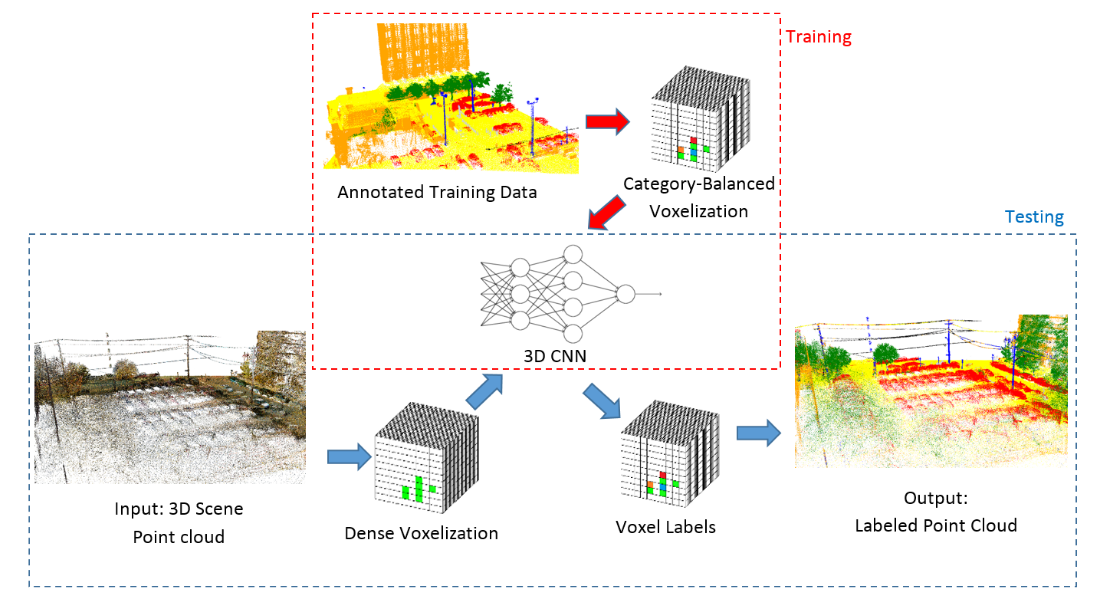}
	\caption{\acs{3D}\acs{CNN} based system presented by Huang \emph{et al.}\cite{Huang2016} for semantic labeling of point clouds. Clouds undergo a dense voxelization process and the \acs{CNN} produces per-voxel labels that are then mapped back to the point cloud. Figure extracted from \cite{Huang2016}.}
	\label{fig:huang3dcnn}
\end{figure}

\acs{3D} geometric data such as point clouds or polygonal meshes are useful representations thanks to their additional dimension which provides methods with rich spatial information that is intuitively useful for segmentation. However, the vast majority of successful deep learning segmentation architectures -- \acp{CNN} in particular -- are not originally engineered to deal with unstructured or irregular inputs such as the aforementioned ones. In order to enable weight sharing and other optimizations in convolutional architectures, most researchers have resorted to 3D voxel grids or projections to transform unstructured and unordered point clouds or meshes into regular representations before feeding them to the networks. For instance, Huang \emph{et al.}\cite{Huang2016} (see Figure \ref{fig:huang3dcnn} take a point cloud and parse it through a dense voxel grid, generating a set of occupancy voxels which are used as input to a 3D \acs{CNN} to produce one label per voxel. They then map back the labels to the point cloud. Although this approach has been applied successfully, it has some disadvantages like quantization, loss of spatial information, and unnecessarily large representations. For that reason, various researchers have focused their efforts on creating deep architectures that are able to directly consume unstructured \acs{3D} point sets or meshes.

\begin{figure*}[!t]
	 \centering
	 \includegraphics[width=0.9\linewidth]{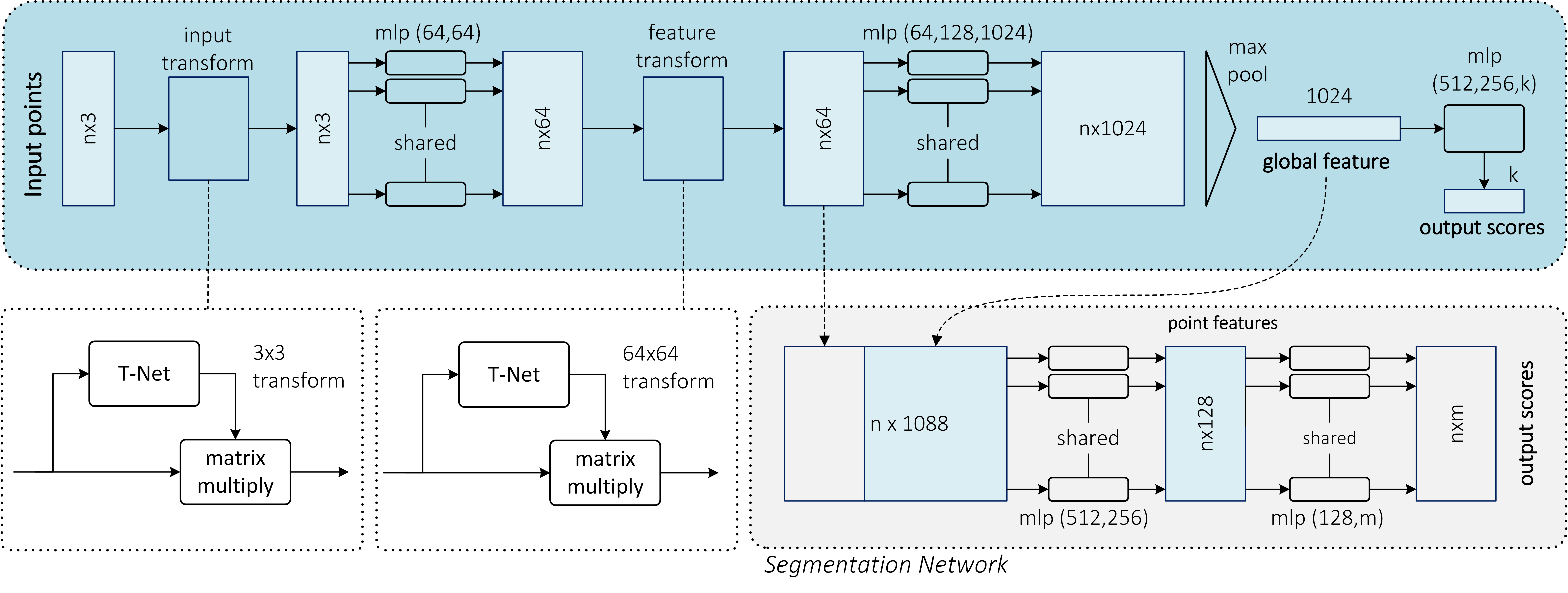}
	 \caption{The PointNet unified architecture for point cloud classification and segmentation. Figure reproduced from \cite{Qi2016}.}
	 \label{fig:pointnetarchitecture}
\end{figure*}

PointNet\cite{Qi2016} is a pioneering work which presents a deep neural network that takes raw point clouds as input, providing a unified architecture for both classification and segmentation. Figure \ref{fig:pointnetarchitecture} shows that two-part network which is able to consume unordered point sets in 3D.

As we can observe, PointNet is a deep network architecture that stands out of the crowd due to the fact that it is based on fully connected layers instead of convolutional ones. The architecture features two subnetworks: one for classification and another for segmentation. The classification subnetwork takes a point cloud and applies a set of transforms and \acp{MLP} to generate features which are then aggregated using max-pooling to generate a global feature which describes the original input cloud. That global feature is classified by another \ac{MLP} to produce output scores for each class. The segmentation subnetwork concatenates the global feature with the per-point features extracted by the classification network and applies another two \acp{MLP} to generate features and produce output scores for each point.

\subsection{Video Sequences}

As we have observed, there has been a significant progress in single-image segmentation. However, when dealing with image sequences, many systems rely on the naïve application of the very same algorithms in a frame-by-frame manner. This approach works, often producing remarkable results. Nevertheless, applying those methods frame by frame is usually non-viable due to computational cost. In addition, those methods completely ignore temporal continuity and coherence cues which might help increase the accuracy of the system while reducing its execution time.

\begin{figure}[!hbt]
	 \centering
	 \includegraphics[width=\linewidth]{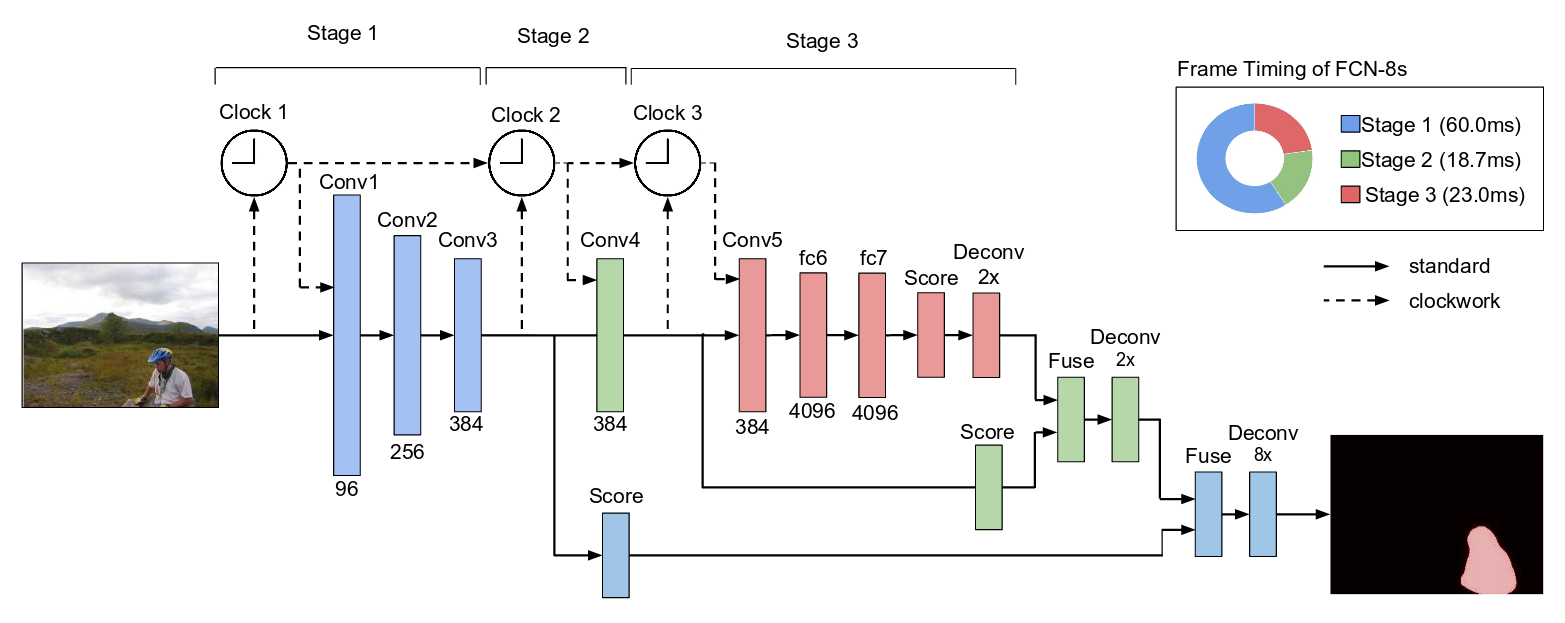}
	 \caption{The clockwork \acs{FCN} with three stages and their corresponding clock rates. Figure extracted from \cite{Shelhamer2016}.}
	 \label{fig:clockworkfcn}
\end{figure}

Arguably, the most remarkable work in this regard is the clockwork \ac{FCN} by Shelhamer \emph{et al.}\cite{Shelhamer2016}. This network is an adaptation of a \ac{FCN} to make use of temporal cues in video to decrease inference time while preserving accuracy. The clockwork approach relies on the following insight: feature velocity -- the temporal rate of change of features in the network -- across frames varies from layer to layer so that features from shallow layers change faster than deep ones. Under that assumption, layers can be grouped into stages, processing them at different update rates depending on their depth. By doing this, deep features can be persisted over frames thanks to their semantic stability, thus saving inference time. Figure \ref{fig:clockworkfcn} shows the network architecture of the clockwork \ac{FCN}.

\begin{figure}[!hbt]
	 \centering
	 \includegraphics[width=\linewidth]{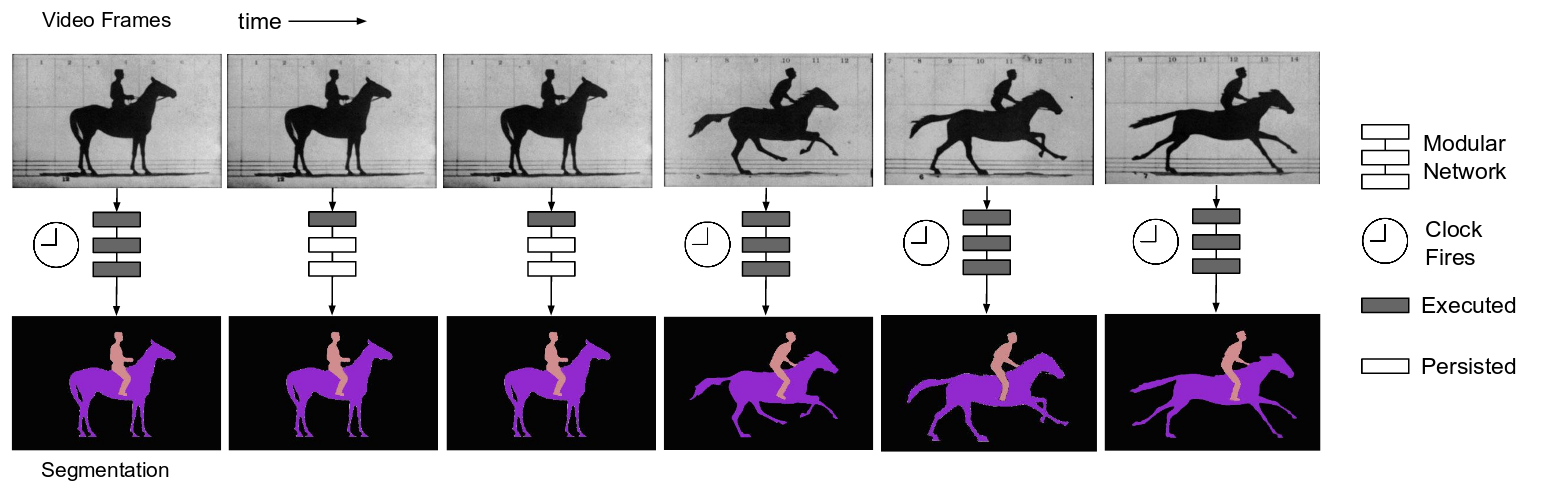}
	 \caption{Adaptive clockwork method proposed by Shelhamer \emph{et al.}\cite{Shelhamer2016}. Extracted features persists during static frames while they are recomputed for dynamic ones. Figure extracted from \cite{Shelhamer2016}.}
	 \label{fig:clockworkfcn-adaptive}
\end{figure}

It is important to remark that the authors propose two kinds of update rates: fixed and adaptive. The fixed schedule just sets a constant time frame for recomputing the features for each stage of the network. The adaptive schedule fires each clock on a data-driven manner, e.g., depending on the amount of motion or semantic change. Figure \ref{fig:clockworkfcn-adaptive} shows an example of this adaptive scheduling.

Zhang \emph{et al.}\cite{Zhang2014} took a different approach and made use of a \acs{3D}\acs{CNN}, which was originally created for learning features from volumes, to learn hierarchical spatio-temporal features from multi-channel inputs such as video clips. In parallel, they over-segment the input clip into supervoxels. Then they use that supervoxel graph and embed the learned features in it. The final segmentation is obtained by applying graph-cut\cite{Boykov2001} on the supervoxel graph.

\begin{figure}[!hbt]
	\centering
	\begin{subfigure}{0.49\linewidth}
		\includegraphics[width=\linewidth]{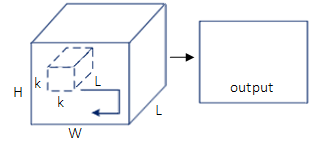}
		\caption{2D Convolution}
		\label{fig:2dvs3dconvolutions:2dconvolution}
	\end{subfigure}
	\hfill
	\begin{subfigure}{0.49\linewidth}
		\includegraphics[width=\linewidth]{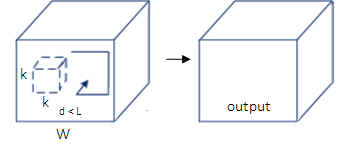}
		\caption{3D Convolution}
		\label{fig:2dvs3dconvolutions:3dconvolution}
	\end{subfigure}
	\caption{Difference between 2D and 3D convolutions applied on a set of frames. (\protect\subref{fig:2dvs3dconvolutions:2dconvolution}) 2D convolutions use the same weights for the whole depth of the stack of frames (multiple channels) and results in a single image. (\protect\subref{fig:2dvs3dconvolutions:3dconvolution}) 3D convolutions use 3D filters and produce a 3D volume as a result of the convolution, thus preserving temporal information of the frame stack.}
	\label{fig:2dvs3dconvolutions}
\end{figure}

Another remarkable method, which builds on the idea of using 3D convolutions, is the deep end-to-end voxel-to-voxel prediction system by Tran \emph{et al.}\cite{Tran2016}. In that work, they make use of the \ac{C3D} network introduced by themselves on a previous work \cite{Tran2015}, and extend it for semantic segmentation by adding deconvolutional layers at the end. Their system works by splitting the input into clips of $16$ frames, performing predictions for each clip separately. Its main contribution is the use of 3D convolutions. Those convolutions make use of \acl{3D} filters which are suitable for spatio-temporal feature learning across multiple channels, in this case frames. Figure \ref{fig:2dvs3dconvolutions} shows the difference between 2D and 3D convolutions applied to multi-channel inputs, proving the usefulness of the 3D ones for video segmentation.

\section{Discussion}
\label{sec:discussion}

In the previous section we reviewed the existing methods from a literary and qualitative point of view, i.e., we did not take any quantitative result into account. In this Section we are going to discuss the very same methods from a numeric standpoint. First of all, we will describe the most popular evaluation metrics that can be used to measure the performance of semantic segmentation systems from three aspects: execution time, memory footprint, and accuracy. Next, we will gather the results of the methods on the most representative datasets using the previously described metrics. After that, we will summarize and draw conclusions about those results. At last, we enumerate possible future research lines that we consider significant for the field.

\subsection{Evaluation Metrics}

For a segmentation system to be useful and actually produce a significant contribution to the field, its performance must be evaluated with rigor. In addition, that evaluation must be performed using standard and well-known metrics that enable fair comparisons with existing methods. Furthermore, many aspects must be evaluated to assert the validity and usefulness of a system: execution time, memory footprint, and accuracy. Depending on the purpose or the context of the system, some metrics might be of more importance than others, i.e., accuracy may be expendable up to a certain point in favor of execution speed for a real-time application. Nevertheless, for the sake of scientific rigor it is of utmost importance to provide all the possible metrics for a proposed method.

\subsubsection{Execution Time}

Speed or runtime is an extremely valuable metric since the vast majority of systems must meet hard requirements on how much time can they spend on the inference pass. In some cases it might be useful to know the time needed for training the system, but it is usually not that significant, unless it is exaggeratedly slow, since it is an offline process. In any case, providing exact timings for the methods can be seen as meaningless since they are extremely dependant on the hardware and the backend implementation, rendering some comparisons pointless.

However, for the sake of reproducibility and in order to help fellow researchers, it is useful to provide timings with a thorough description of the hardware in which the system was executed on, as well as the conditions for the benchmark. If done properly, that can help others estimate if the method is useful or not for the application as well as perform fair comparisons under the same conditions to check which are the fastest methods.

\subsubsection{Memory Footprint}

Memory usage is another important factor for segmentation methods. Although it is arguably less constraining than execution time -- scaling memory capacity is usually feasible -- it can also be a limiting element. In some situations, such as onboard chips for robotic platforms, memory is not as abundant as in a high-performance server. Even high-end \acp{GPU}, which are commonly used to accelerate deep networks, do not pack a copious amount of memory. In this regard, and considering the same implementation-dependent aspects as with runtime, documenting the peak and average memory footprint of a method with a complete description of the execution conditions can be extraordinarily helpful.

\subsubsection{Accuracy}

Many evaluation criteria have been proposed and are frequently used to assess the accuracy of any kind of technique for semantic segmentation. Those metrics are usually variations on pixel accuracy and \ac{IoU}. We report the most popular metrics for semantic segmentation that are currently used to measure how per-pixel labeling methods perform on this task. For the sake of the explanation, we remark the following notation details: we assume a total of $k+1$ classes (from $L_0$ to $L_{k}$ including a void class or background) and $p_{ij}$ is the amount of pixels of class $i$ inferred to belong to class $j$. In other words, $p_{ii}$ represents the number of true positives, while $p_{ij}$ and $p_{ji}$ are usually interpreted as false positives and false negatives respectively (although either of them can be the sum of both false positives and false negatives)..

\begin{itemize}
	\item \textbf{\ac{PA}}: it is the simplest metric, simply computing a ratio between the amount of properly classified pixels and the total number of them.
		\begin{align*}
			PA = \displaystyle\frac{\displaystyle\sum_{i=0}^k p_{ii}}{\displaystyle\sum_{i=0}^k\displaystyle\sum_{j=0}^k p_{ij}}
		\end{align*}
	\item \textbf{\ac{MPA}}: a slightly improved \ac{PA} in which the ratio of correct pixels is computed in a per-class basis and then averaged over the total number of classes.
		\begin{align*}
			MPA = \displaystyle\frac{1}{k+1}\displaystyle\sum_{i=0}^k \displaystyle\frac{p_{ii}}{\displaystyle\sum_{j=0}^k p_{ij}}
		\end{align*}
	\item \textbf{\ac{MIoU}}: this is the standard metric for segmentation purposes. It computes a ratio between the intersection and the union of two sets, in our case the ground truth and our predicted segmentation. That ratio can be reformulated as the number of true positives (intersection) over the sum of true positives, false negatives, and false positives (union). That \ac{IoU} is computed on a per-class basis and then averaged.
		\begin{align*}
			MIoU = \displaystyle\frac{1}{k+1}\displaystyle\sum_{i=0}^k \displaystyle\frac{p_{ii}}{\displaystyle\sum_{j=0}^k p_{ij} + \displaystyle\sum_{j=0}^k p_{ji} - p_{ii}}
		\end{align*}
	\item \textbf{\ac{FWIoU}}: it is an improved over the raw \ac{MIoU} which weights each class importance depending on their appearance frequency.
		\begin{align*}
			FWIoU = \displaystyle\frac{1}{\displaystyle\sum_{i=0}^k \displaystyle\sum_{j=0}^k p_{ij}} \displaystyle\sum_{i=0}^k \displaystyle\frac{\displaystyle\sum_{j=0}^k p_{ij} p_{ii}}{\displaystyle\sum_{j=0}^k p_{ij} + \displaystyle\sum_{j=0}^k p_{ji} - p_{ii}}
		\end{align*}
\end{itemize}

Of all metrics described above, the \ac{MIoU} stands out of the crowd as the most used metric due to its representativeness and simplicity. Most challenges and researchers make use of that metric to report their results.

\subsection{Results}

As we stated before, Section \ref{sec:methods} provided a functional description of the reviewed methods according to their targets. Now we gathered all the quantitative results for those methods as stated by their authors in their corresponding papers. These results are organized into three parts depending on the input data used by the methods: \acs{2D} \acs{RGB} or \acs{2.5D} \acs{RGB-D} images, volumetric \acs{3D}, or video sequences.

The most used datasets have been selected for that purpose. It is important to remark the heterogeneity of the papers in the field when reporting results. Although most of them try to evaluate their methods in standard datasets and provide enough information to reproduce their results, also expressed in widely known metrics, many others fail to do so. That leads to a situation in which it is hard or even impossible to fairly compare methods.

Furthermore, we also came across the fact few authors provide information about other metrics rather than accuracy. Despite the importance of other metrics, most of the papers do not include any data about execution time nor memory footprint. In some cases that information is provided, but no reproducibility information is given so it is impossible to know the setup that produced those results which are of no use.

\subsubsection{\acs{RGB}}

For the single \acs{2D} image category we have selected seven datasets: PASCAL \acs{VOC}2012, PASCAL Context, PASCAL Person-Part, CamVid, CityScapes, Stanford Background, and SiftFlow. That selection accounts for a wide range of situations and targets.

The first, and arguably the most important dataset, in which the vast majority of methods are evaluated is PASCAL \acs{VOC}-2012. Table \ref{table:pascal-voc-2012} shows the results of those reviewed methods which provide accuracy results on the PASCAL \acs{VOC}-2012 test set. This set of results shows a clear improvement trend from the firs proposed methods (SegNet and the original \acs{FCN}) to the most complex models such as \acs{CRF}as\acs{RNN} and the winner (DeepLab) with $79.70$ \acs{IoU}.

\begin{table}[!htb]
	\centering
	\caption{Performance results on PASCAL \acs{VOC}-2012.}
	\label{table:pascal-voc-2012}
		\begin{tabular}{|c|c|c|}
			\hline
			\# & Method & Accuracy (\acs{IoU})\\
			\hline
			1 & DeepLab\cite{Chen2016} & $79.70$\\
			2 & Dilation\cite{Yu2015} & $75.30$\\
			3 & CRFasRNN\cite{Zheng2015} & $74.70$\\
			4 & ParseNet\cite{Liu2015} & $69.80$\\
			5 & \acs{FCN}-8s\cite{Long2015} & $67.20$\\
			6 & Multi-scale-CNN-Eigen\cite{Eigen2015} & $62.60$\\
			7 & Bayesian SegNet\cite{Kendall2015} & $60.50$\\
			\hline
		\end{tabular}
\end{table}

Apart from the widely known \acs{VOC} we also collected metrics of its Context counterpart. Table \ref{table:pascal-context} shows those results in which DeepLab is again the top scorer ($45.70$ \acs{IoU}).

\begin{table}[!htb]
	\centering
	\caption{Performance results on PASCAL-Context.}
	\label{table:pascal-context}
		\begin{tabular}{|c|c|c|}
			\hline
			\# & Method & Accuracy (\acs{IoU})\\
			\hline
			1 & DeepLab\cite{Chen2016} & $45.70$\\
			2 & CRFasRNN\cite{Zheng2015} & $39.28$\\
			3 & \acs{FCN}-8s\cite{Long2015} & $39.10$\\
			\hline
		\end{tabular}
\end{table}

In addition, we also took into account the PASCAL Part dataset, whose results are shown in Table \ref{table:pascal-person-part}. In this case, the only analyzed method that provided metrics for this dataset is DeepLab which achieved a $64.94$ \acs{IoU}.

\begin{table}[!htb]
	\centering
	\caption{Performance results on PASCAL-Person-Part.}
	\label{table:pascal-person-part}
		\begin{tabular}{|c|c|c|}
			\hline
			\# & Method & Accuracy (\acs{IoU})\\
			\hline
			1 & DeepLab\cite{Chen2016} & $64.94$\\
			\hline
		\end{tabular}
\end{table}

Moving from a general-purpose dataset such as PASCAL \acs{VOC}, we also gathered results for two of the most important urban driving databases. Table \ref{table:camvid} shows the results of those methods which provide accuracy metrics for the CamVid dataset. In this case, an \acs{RNN}-based approach (DAG-\acs{RNN}) is the top one with a $91.60$ \acs{IoU}.

\begin{table}[!htb]
	\centering
	\caption{Performance results on CamVid.}
	\label{table:camvid}
		\begin{tabular}{|c|c|c|}
			\hline
			\# & Method & Accuracy (\acs{IoU})\\
			\hline
			1 & DAG-RNN \cite{Shuai2015} & $91.60$\\
			2 & Bayesian SegNet\cite{Kendall2015} & $63.10$\\
			3 & SegNet\cite{Badrinarayanan2015} & $60.10$\\
			4 & ReSeg \cite{Visin2016} & $58.80$\\
			5 & ENet\cite{Paszke2016} & $55.60$\\
			\hline
		\end{tabular}
\end{table}

Table \ref{table:cityscapes} shows the results on a more challenging and currently more in use database: CityScapes. The trend on this dataset is similar to the one with PASCAL \acs{VOC} with DeepLab leading with a $70.40$ \acs{IoU}.

\begin{table}[!htb]
	\centering
	\caption{Performance results on CityScapes.}
	\label{table:cityscapes}
		\begin{tabular}{|c|c|c|}
			\hline
			\# & Method & Accuracy (\acs{IoU})\\
			\hline
			1 & DeepLab\cite{Chen2016} & $70.40$\\
			2 & Dilation10\cite{Yu2015} & $67.10$\\
			3 & \acs{FCN}-8s\cite{Long2015} & $65.30$\\
			4 & CRFasRNN\cite{Zheng2015} & $62.50$\\
			5 & ENet\cite{Paszke2016} & $58.30$\\
			\hline
		\end{tabular}
\end{table}

Table \ref{table:stanford-background} shows the results of various recurrent networks on the Stanford Background dataset. The winner, r\acs{CNN}, achieves a maximum accuracy of $80.20$ \acs{IoU}.

\begin{table}[!htb]
	\centering
	\caption{Performance results on Stanford Background.}
	\label{table:stanford-background}
		\begin{tabular}{|c|c|c|}
			\hline
			\# & Method & Accuracy (\acs{IoU}) \\
			\hline
			1 & rCNN\cite{Pinheiro2014} & $80.20$ \\
			2 & 2D-LSTM\cite{Byeon2015} & $78.56$ \\
			\hline
		\end{tabular}
\end{table}

At last, results for another popular dataset such as SiftFlow are shown in Table \ref{table:siftflow}. This dataset is also dominated by recurrent methods. In particular DAG-\acs{RNN} is the top scorer with $85.30$ \acs{IoU}.

\begin{table}[!htb]
	\centering
	\caption{Performance results on SiftFlow.}
	\label{table:siftflow}
		\begin{tabular}{|c|c|c|}
			\hline
			\# & Method & Accuracy (\acs{IoU})\\
			\hline
			1 & DAG-RNN\cite{Shuai2015} & $85.30$ \\
			2 & rCNN\cite{Pinheiro2014} & $77.70$ \\
			3 & 2D-LSTM\cite{Byeon2015} & $70.11$ \\
			\hline
		\end{tabular}
\end{table}

\subsubsection{\acs{2.5D}}

Regarding the \acs{2.5D} category, i.e., datasets which also include depth information apart from the typical \acs{RGB} channels, we have selected three of them for the analysis: SUN-\acs{RGB-D} and NYUDv2. Table \ref{table:sunrgbd} shows the results for SUN-\acs{RGB-D} that are only provided by \acs{LSTM-CF}, which achieves $48.10$ \acs{IoU}.

\begin{table}[!htb]
	\centering
	\caption{Performance results on SUN-\acs{RGB-D}.}
	\label{table:sunrgbd}
	\begin{tabular}{|c|c|c|}
		\hline
		\# & Method & Accuracy (\acs{IoU})\\
		\hline
		1 & \acs{LSTM-CF}\cite{Li2016b} & $48.10$ \\
		\hline
	\end{tabular}
\end{table}

Table \ref{table:nyudv2} shows the results for NYUDv2 which are exclusive too for \acs{LSTM-CF}. That method reaches $49.40$ \acs{IoU}.

\begin{table}[!htb]
	\centering
	\caption{Performance results on NYUDv2.}
	\label{table:nyudv2}
	\begin{tabular}{|c|c|c|}
		\hline
		\# & Method & Accuracy (\acs{IoU})\\
		\hline
		1 & \acs{LSTM-CF}\cite{Li2016b} & $49.40$ \\
		\hline
	\end{tabular}
\end{table}

At last, Table \ref{table:sun3d} gathers results for the last \acs{2.5D} dataset: SUN-\acs{3D}. Again, \acs{LSTM-CF} is the only one which provides information for that database, in this case a $58.50$ accuracy.

\begin{table}[!htb]
	\centering
	\caption{Performance results on SUN3D.}
	\label{table:sun3d}
	\begin{tabular}{|c|c|c|}
		\hline
		\# & Method & Accuracy (\acs{IoU})\\
		\hline
		1 & \acs{LSTM-CF}\cite{Li2016b} & $58.50$ \\
		\hline
	\end{tabular}
\end{table}

\subsubsection{\acs{3D}}

Two \acs{3D} datasets have been chosen for this discussion: ShapeNet Part and Stanford-\acs{2D}-\acs{3D}-S. In both cases, only one of the analyzed methods actually scored on them. It is the case of PointNet which achieved $83.80$ and $47.71$ \acs{IoU} on ShapeNet Part (Table \ref{table:shapenet-part}) and Stanford-\acs{2D}-\acs{3D}-S (Table \ref{table:stanford-2d-3d-s}) respectively.

\begin{table}[!htb]
	\centering
	\caption{Performance results on ShapeNet Part.}
	\label{table:shapenet-part}
		\begin{tabular}{|c|c|c|}
			\hline
			\# & Method & Accuracy (\acs{IoU})\\
			\hline
			1 & PointNet\cite{Qi2016} & $83.70$\\
			\hline
		\end{tabular}
\end{table}

\begin{table}[!htb]
	\centering
	\caption{Performance results on Stanford 2D-3D-S.}
	\label{table:stanford-2d-3d-s}
		\begin{tabular}{|c|c|c|}
			\hline
			\# & Method & Accuracy (\acs{IoU})\\
			\hline
			1 & PointNet\cite{Qi2016} & $47.71$\\
			\hline
		\end{tabular}
\end{table}

\subsubsection{Sequences}

The last category included in this discussion is video or sequences. For that part we gathered results for two datasets which are suitable for sequence segmentation: CityScapes and YouTube-Objects. Only one of the reviewed methods for video segmentation provides quantitative results on those datasets: Clockwork Convnet. That method reaches $64.40$ \acs{IoU} on CityScapes (Table \ref{table:cityscapes-seq}), and $68.50$ on YouTube-Objects (Table \ref{table:youtube-objects}).

\begin{table}[!htb]
	\centering
	\caption{Performance results on Cityscapes.}
	\label{table:cityscapes-seq}
		\begin{tabular}{|c|c|c|}
			\hline
			\# & Method & Accuracy (\acs{IoU})\\
			\hline
			1 & Clockwork Convnet\cite{Shelhamer2016} & $64.40$\\
			\hline
		\end{tabular}
\end{table}

\begin{table}[!htb]
	\centering
	\caption{Performance results on Youtube-Objects.}
	\label{table:youtube-objects}
		\begin{tabular}{|c|c|c|}
			\hline
			\# & Method & Accuracy (\acs{IoU})\\
			\hline
			1 & Clockwork Convnet\cite{Shelhamer2016} & $68.50$\\
			\hline
		\end{tabular}
\end{table}

\subsection{Summary}

In light of the results, we can draw various conclusions. The most important of them is related to reproducibility. As we have observed, many methods report results on non-standard datasets or they are not even tested at all. That makes comparisons impossible. Furthermore, some of them do not describe the setup for the experimentation or do not provide the source code for the implementation, thus significantly hurting reproducibility. Methods should report their results on standard datasets, exhaustively describe the training procedure, and also make their models and weights publicly available to enable progress.

Another important fact discovered thanks to this study is the lack of information about other metrics such as execution time and memory footprint. Almost no paper reports this kind of information, and those who do suffer from the reproducibility issues mentioned before. This void is due to the fact that most methods focus on accuracy without any concern about time or space. However, it is important to think about where are those methods being applied. In practice, most of them will end up running on embedded devices, e.g., self-driving cars, drones, or robots, which are fairly limited from both sides: computational power and memory.

Regarding the results themselves, we can conclude that DeepLab is the most solid method which outperforms the rest on almost every single \acs{RGB} images dataset by a significant margin. The \acs{2.5D} or multimodal datasets are dominated by recurrent networks such as \acs{LSTM-CF}. \acs{3D} data segmentation still has a long way to go with PointNet paving the way for future research on dealing with unordered point clouds without any kind of preprocessing or discretization. Finally, dealing with video sequences is another green area with no clear direction, but Clockwork Convnets are the most promising approach thanks to their efficiency and accuracy duality. \acs{3D} convolutions are worth remarking due to their power and flexibility to process multichannel inputs, making them successful at capturing both spatial and temporal information.

\subsection{Future Research Directions}

Based on the reviewed research, which marks the state of the art of the field, we present a list of future research directions that would be interesting to pursue.

\begin{itemize}
	\item \emph{3D datasets}: methods that make full use of \acs{3D} information are starting to rise but, even if new proposals and techniques are engineered, they still lack one of the most important components: data. There is a strong need for large-scale datasets for \acs{3D} semantic segmentation, which are harder to create than their lower dimensional counterparts. Although there are already some promising works, there is still room for more, better, and varied data. It is important to remark the importance of real-world 3D data since most of the already existing works are synthetic databases. A proof of the importance of \acs{3D} is the fact that the \acs{ILSVRC} will feature \acs{3D} data in 2018.
	\item \emph{Sequence datasets}: the same lack of large-scale data that hinders progress on \acs{3D} segmentation also impacts video segmentation. There are only a few datasets that are sequence-based and thus helpful for developing methods which take advantage of temporal information. Bringing up more high-quality data from this nature, either \acs{2D} or \acs{3D}, will unlock new research lines without any doubt.
	\item \emph{Point cloud segmentation using \acp{GCN}}: as we already mentioned, dealing with \acs{3D} data such as point clouds poses an unsolved challenge. Due to its unordered and unstructured nature, traditional architectures such as \acp{CNN} cannot be applied unless some sort of discretization process is applied to structure it. One promising line of research aims to treat point clouds as graphs and apply convolutions over them \cite{Henaff2015} \cite{Kipf2016} \cite{Niepert2016}. This has the advantage of preserving spatial cues in every dimension without quantizing data.
	\item \emph{Context knowledge}: while \acp{FCN} are a consolidated approach for semantic segmentation, they lack several features such as context modelling that help increasing accuracy. The reformulation of CRFs as RNNs to create end-to-end solutions seems to be a promising direction to improve results on real-life data. Multi-scale and feature fusion approaches have also shown remarkable progress. In general, all those works represent important steps towards achieving the ultimate goal, but there are some problems that still require more research.
	\item \emph{Real-time segmentation}: In many applications, precision is important; however, it is also crucial that these implementations are able to cope with common camera frame rates (at least 25 frames per second). Most of the current methods are far from that framerate, e.g., \acs{FCN}-8s takes roughly $100$ ms to process a low-resolution PASCAL \acs{VOC} image whilst \acs{CRF}as\acs{RNN} needs more than $500$ ms. Therefore, during the next years, we expect a stream of works coming out, focusing more on real-time constraints. These future works will have to find a trade-off between accuracy and runtime.
	\item \emph{Memory:} some platforms are bounded by hard memory constraints. Segmentation networks usually do need significant amounts of memory to be executed for both inference and training. In order to fit them in some devices, networks must be simplified. While this can be easily accomplished by reducing their complexity (often trading it for accuracy), another approaches can be taken. Pruning is a promising research line that aims to simplify a network, making it lightweight while keeping the knowledge, and thus the accuracy, of the original network architecture \cite{Anwar2015}\cite{Han2015}\cite{Molchanov2016}.
	\item \emph{Temporal coherency on sequences:} some methods have addressed video or sequence segmentation but either taking advantage of that temporal cues to increase accuracy or efficiency. However, none of them have explicitly tackled the coherency problem. For a segmentation system to work on video streams it is important, not only to produce good results frame by frame, but also make them coherent through the whole clip without producing artifacts by smoothing predicted per-pixel labels along the sequence.
	\item \emph{Multi-view integration:} Use of multiple views in recently proposed segmentation works is mostly limited to RGB-D cameras and in particular focused on single-object segmentation.
\end{itemize}

\section{Conclusion}
\label{sec:conclusion}

To the best of our knowledge, this is the first review paper in the literature which focuses on semantic segmentation using deep learning. In comparison with other surveys, this paper is devoted to such a rising topic as deep learning, covering the most advanced and recent work on that front. We formulated the semantic segmentation problem and provided the reader with the necessary background knowledge about deep learning for such task. We covered the contemporary literature of datasets and methods, providing a comprehensive survey of $28$ datasets and $27$ methods. Datasets were carefully described, stating their purposes and characteristics so that researchers can easily pick the one that best suits their needs. Methods were surveyed from two perspectives: contributions and raw results, i.e., accuracy. We also presented a comparative summary of the datasets and methods in tabular forms, classifying them according to various criteria. In the end, we discussed the results and provided useful insight in shape of future research directions and open problems in the field. In conclusion, semantic segmentation has been approached with many success stories but still remains an open problem whose solution would prove really useful for a wide set of real-world applications. Furthermore, deep learning has proved to be extremely powerful to tackle this problem so we can expect a flurry of innovation and spawns of research lines in the upcoming years.

\ifCLASSOPTIONcompsoc
  \section*{Acknowledgments}
\else
  \section*{Acknowledgment}
\fi

This work has been funded by the Spanish Government TIN2016-76515-R grant for the COMBAHO project, supported with Feder funds. It has also been supported by a Spanish national grant for PhD studies FPU15/04516. In addition, it was also funded by the grant \emph{Ayudas para Estudios de Máster e Iniciación a la Investigación} from the University of Alicante.

\ifCLASSOPTIONcaptionsoff
  \newpage
\fi



%

\bibliographystyle{IEEEtran}
\bibliography{references}

%

	\begin{IEEEbiography}[{\includegraphics[width=1in,height=1.25in,clip,keepaspectratio]{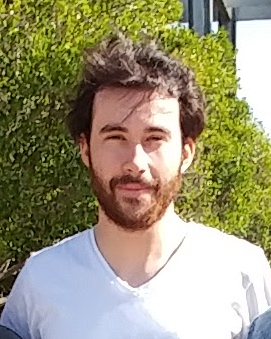}}]{Alberto Garcia-Garcia} is a PhD Student (Machine Learning and Computer Vision) at the University of Alicante. He received his Master's Degree (Automation and Robotics) and his Bachelor's Degree (Computer Engineering) from the same institution in June 2015 and June 2016 respectively. His main research interests include deep learning (specially convolutional neural networks), 3D computer vision, and parallel computing on GPUs. He was an intern at Jülich Supercomputing Center, and at NVIDIA working jointly with the Camera/Solutions engineering team and the Mobile Visual Computing group from NVIDIA Research. He is also a member of European Networks such as HiPEAC and IV\&L.
\end{IEEEbiography}

\begin{IEEEbiography}[{\includegraphics[width=1in,height=1.25in,clip,keepaspectratio]{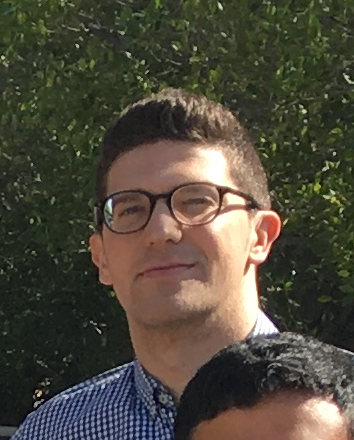}}]{Sergio Orts-Escolano} received a BSc, MSc and PhD in Computer Science from the University of Alicante (Spain) in 2008, 2010 and 2014 respectively. He is currently an assistant professor in the Department of Computer Science and Artificial Intelligence at the University of Alicante. Previously he was a researcher at Microsoft Research where he was one of the leading members of the Holoportation project (virtual 3D teleportation in real-time). His research interests include computer vision, 3D sensing, real-time computing, GPU computing, and deep learning.  He has authored +50 publications in journals and top conferences like CVPR, SIGGRAPH, 3DV, BMVC, Neurocomputing, Neural Networks, Applied Soft Computing, etcetera. He is also member of European Networks like HiPEAC and Eucog. 
\end{IEEEbiography}

\begin{IEEEbiography}[{\includegraphics[width=1in,height=1.25in,clip,keepaspectratio]{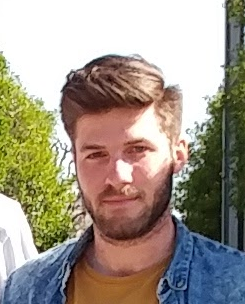}}]{Sergiu-Ovidiu Oprea}
is a MSc Student (Automation and Robotics) at University of Alicante. He received his Bachelor's Degree (Computer Engineering) from the same institution in June 2015. His main research interests include deep learning (specially recurrent neural networks), 3D computer vision, parallel computing on GPUs, and computer graphics. He is also member of European Networks like HiPEAC.
\end{IEEEbiography}

\begin{IEEEbiography}[{\includegraphics[width=1in,height=1.25in,clip,keepaspectratio]{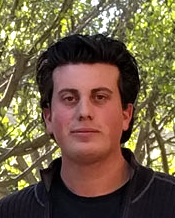}}]{Victor Villena-Martinez}
is a PhD Student at the University of Alicante. He received his Master's Degree in Automation and Robotics in June 2016 and his Bachelor's Degree in Computer engineering in June 2015. He has collaborated in the project "Acquisition and modeling of growing plants" (GV/2013/005). His main research is focused on the calibration of RGB-D devices and the reconstruction of the human body using the same devices.
\end{IEEEbiography}


\begin{IEEEbiography}[{\includegraphics[width=1in,height=1.25in,clip,keepaspectratio]{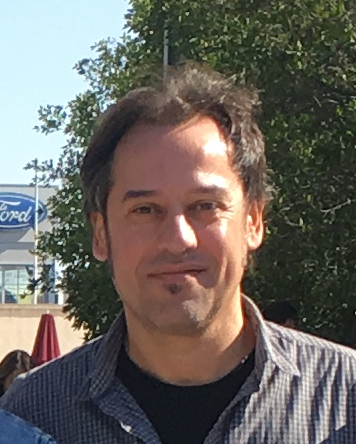}}]{Jose Garcia-Rodriguez}
	received his Ph.D. degree, with specialization in Computer Vision and Neural Networks, from the University of Alicante (Spain). He is currently Associate Professor at the Department of Computer Technology of the University of Alicante. His research areas of interest include: computer vision, computational intelligence, machine learning, pattern recognition, robotics, man-machine interfaces, ambient intelligence, computational chemistry, and parallel and multicore architectures. He has authored +100 publications in journals and top conferences and revised papers for several journals like Journal of Machine Learning Research, Computational intelligence, Neurocomputing, Neural Networks, Applied Softcomputing, Image Vision and Computing, Journal of Computer Mathematics, IET on Image Processing, SPIE Optical Engineering and many others, chairing sessions in the last decade for WCCI/IJCNN and participating in program committees of several conferences including IJCNN, ICRA, ICANN, IWANN, IWINAC KES, ICDP and many others. He is also member of European Networks of Excellence and COST actions like Eucog, HIPEAC, AAPELE or I\&VL and director or the GPU Research Center at University of Alicante and Phd program in Computer Science.
\end{IEEEbiography}




\end{document}